\documentclass[11pt,draftcls,onecolumn]{IEEEtran}
\usepackage{epsfig,colordvi,pxfonts,amsfonts}

\title{Colour Image Segmentation by the Vector-valued Allen-Cahn Phase-field Model: a Multigrid Solution}

\author{David A Kay
\thanks{Oxford Computing Laboratory, Wolfson Building, Parks Road, Oxford, OX1 3QD, UK. David.Kay@comlab.ox.ac.uk Tel: 0044 (0)1865 610814}
\and $^*$ Alessandro Tomasi
\thanks{Department of Mathematics, University of Sussex, Falmer, Brighton, BN1 9RF, UK. keug1@sussex.ac.uk Tel: 0044 (0)1273 873450. Corresponding author.}
}
\date{}

\begin{document}
\maketitle

\begin{abstract}

We propose a new method for the numerical solution of a PDE-driven model for
colour image segmentation and give numerical examples of the results. The method combines the
vector-valued Allen-Cahn phase field equation with initial data fitting terms. This method is known to be closely
related to the Mumford-Shah problem and the level set segmentation by Chan and Vese. Our
numerical solution is performed using a multigrid splitting of a finite element space,
thereby producing an efficient and robust method for the segmentation of large images.


This work has been submitted to the IEEE for possible
publication. Copyright may be transferred without notice, after
which this version may no longer be accessible.

\end{abstract}

\section{Introduction}

The Mumford-Shah functional was first proposed in \cite{Mumford_Shah} as a general way to pose
the problem of image segmentation. Reviews can be found for example in Petitot \cite{Petitot}, and Fusco \cite{Fusco}.
The generality of its statement has led to several methods of solution; in particular, the model
sometimes referred to as the reduced Mumford-Shah was solved by the level set method by Chan and Vese
\cite{Chan_Vese_98}, based on a previous paper on the motion of multiphase junctions tracked by the level set method by
Zhao, Chan, Merriman and Osher \cite{Zhao_Chan_Merriman_Osher_96}, and subsequently extended in Chan and Vese
\cite{Chan_Vese_99}, \cite{Chan_Vese_00}, \cite{Chan_Vese_01}, Vese \cite{Vese_02}, Chan, Shen and Vese
\cite{Chan_Shen_Vese_02}. Due to the extent of their work, it is also often referred to as the Chan-Vese model.
The level set is used to track the boundaries of objects and should converge to the set of contours in the image.

Esedo\={g}lu and Tsai \cite{Esedoglu_Tsai} proposed the Allen-Cahn equation, also known as the phase field model,
as a method of solution to the reduced Mumford-Shah problem when used in conjunction with the Chan-Vese fitting terms.
The Allen-Cahn equation has been used in the context of image processing by Bene\v{s}, Chalupeck\'{y}
and Mikula \cite{Benes_Chalupecky_Mikula}, who first convolve the image with a Gaussian smoothing kernel
to eliminate noise, and then use it as an anisotropic gradient filter
based on the proposals by Perona and Malik \cite{Perona_Malik}. Our approach differs significantly in that we
use no smoothing kernel and we modify the energy functional by adding fitting terms; moreover, we use the vector-valued
formulation of the Allen-Cahn equation due to Garcke, Nestler and Stoth \cite{Garcke_Nestler_Stoth_97}.

A different phase transition model (Modica-Mortola) was also recently used by Jung, Kang and Shen
\cite{Jung_Kang_Shen_06}, and although the model and numerical
method of solution therein differs from this work, it is in many ways similar in the fundamental approach of
adapting a phase transition model to solve the Mumford-Shah problem, and in the results obtained.

Our computations are solved by a multigrid algorithm which falls into the
category of Successive Subspace Corrections (see Xu \cite{Xu_92}, \cite{Xu_01}).
This was successfully applied to the vector-valued Allen-Cahn equation in Kornhuber,
\cite{Kornhuber_03}, Kornhuber and Krause \cite{Kornhuber_Krause_03}, \cite{Kornhuber_Krause_06} with a
small variation (see Kornhuber and Krause \cite{Kornhuber_94}).

In section \ref{summary_intro} we briefly introduce and summarise previous directly relevant work leading up
to section \ref{section_VV_minimisation}, in which we formally introduce our own formulation and 
show how the minimisation of our functional leads to the
desired system of PDEs; in section \ref{section_discretisation_and_numerics} we discretise the system
and introduce the numerical method of solution, and
in section \ref{section_implementation} we present a few practical aspects of implementation together with examples.

\section{Image segmentation by the Allen-Cahn equation}
\label{summary_intro}

\subsection{Relation to the Mumford-Shah functional}

Given an image $I \in \Omega \subset \mathbb{R}^2$, the Mumford-Shah method seeks to
partition the domain $\Omega$ into several subdomains $\Omega_i$ separated by a set $K$ of boundaries,
also known as edges or discontinuities. This segmentation takes the form of
piecewise smooth functions $u \in \Omega - K$ which are discontinuous on $K$; the method of selection from all
possible functions $u$ is minimisation of an energy functional
\begin{equation}
MS(u) = \int_{\Omega - K}|\nabla u|^2 ~dx + \mu \int_{K} d\sigma + \lambda \int_{\Omega} (u-I)^2 ~dx
\label{MS}
\end{equation}
where $\mu, \lambda$ are positive constants.

The first term minimises the variation of $u$ and promotes its smoothness, the second term minimises
the length of interfaces and determines the boundaries between $\Omega_i$, and the third term, sometimes referred to
as the fidelity or fitting term, minimises the variation between $u$ and $I$.

As noted in Petitot \cite{Petitot}, the coefficients $\mu$ and $\lambda$ define several scales of the problem: low $\mu$
leads to fine-grain segmentation, high $\mu$ to coarse-grain results.
Sensitivity to contrast is measured by $(4 \lambda^2 \mu)^{1/4}$ and robustness to noise depends on $\lambda \mu$.

Many variations on this theme have been proposed since its first formulation. Mumford and Shah themselves pointed out
that a reduced form of the problem, referred to as the minimal partition problem, is the restriction of $u$ to
piecewise constant functions, i.e. $u=c_i$ with each $c_i$ a constant on each connected region $\Omega_i$.
The minimising values are then clearly the averages of $I$ across each region.

In the level set case, using by way of example a single function $\phi$ to segment an image containing only
one object against a background, the Chan-Vese functional introduced in \cite{Chan_Vese_98} replaces the fidelity
term in $MS(u)$ by two fitting terms,
\begin{equation}
F_1(\phi)+F_2(\phi)=\int_{inside(\phi=0)}|I-c_1|^2 ~x + \int_{outside(\phi=0)}|I-c_2|^2 ~dx
\label{fidelity}
\end{equation}
where $c_1$, $c_2$ are the average of $I$ inside and outside $\phi=0$, respectively. Considering for a moment the ideal
situation in which the image contains only one object, i.e. is split into two regions of roughly constant value
clearly separated by a gradient boundary, these two terms are clearly minimised when the set
$\phi=0$ coincides with the contour of the object, i.e. the set $K$.

\subsection{A phase-field formulation}

The Allen-Cahn PDE was introduced in \cite{Allen_Cahn} to model the domain coarsening occurring after a phase transition.
It follows the evolution of a function $u(x)$ known as the \textit{order parameter}, which smoothly varies
between the values of $0$ and $1$ across an interface\footnote{The original model was defined
on the interval $[-1,1]$, but it is convenient to consider $[0,1]$ for our purposes, without loss of generality.}
to represent which parts of the material are in one phase or another. It is obtained by minimising the following energy functional:
\begin{equation}
AC(u)=\int_{\Omega}\epsilon|\nabla u|^2 + \frac{1}{4\epsilon}\Psi(u)~dx
\label{AC}
\end{equation}
The function $\Psi(u)$ represents a potential that attains minimal values at the two extreme values of $u$. This is not in general
a convex function, nor is it necessarily smooth. Esedo\={g}lu and Tsai \cite{Esedoglu_Tsai}, for example, chose the
quartic double-well form of the potential, $\Psi(u) = u^2(1-u)^2$.

Comparing the first two terms in (\ref{MS}) and (\ref{AC}), the first term is identical up to a scaling constant,
while the role of the second term is quite similar since the parameter $\epsilon$ is directly related
to the width of interfaces between phases; it is well known that minimising $\Psi(u)$ as above reduces both
the width and length of boundaries. In this paper, we examine the results of extending (\ref{AC}) to its
vector-valued formulation and combining it with fitting terms such as those in (\ref{fidelity}).

It is reasonable to suggest that the results obtained by a level set method and a phase-field method should be closely comparable
because both are known to be equivalent to curve motion by mean curvature; for an overview, see for example \cite{Elliott_97}.

The method of solution described in \cite{Esedoglu_Tsai} follows the MBO thresholding scheme by
Merriman, Bence and Osher \cite{Merriman_Bence_Osher_92}, 
\cite{Merriman_Bence_Osher_94}, which assigns to the order parameter either one or the other extremal value at each step;
we propose to use the formulation known as the \textit{double obstacle} instead, in which, $\Psi$ takes the form
\[
\Psi(u) = \Phi(u) + Q(u),
\]
where
\[
\Phi(u)= \left\{ \begin{array}{lc}
0 & u \in [0,1] \\
+\infty & u \not\in [0,1]
\end{array} \right.
\]
is known as the \textit{indicator function} on the set $[0,1]$, and $Q(u)$ is the concave quadratic
\[
Q(u) = u(1-u).
\]

\subsection{A modified vector-valued Allen-Cahn equation}
\label{section_VV_minimisation}

Our primary objective is to achieve a fast, robust image segmentation method.
Given a domain $\Omega \subset \mathbb{R}^2$ and an image $I \colon \Omega \rightarrow \mathbb{R}^c$ with $c$
data channels or colours, we follow the motion of a function
$u$ with several ($N$) components that we
want to adapt to the significant features of $I$. We propose to do this by solving a
system of PDEs on a finite-element space by a multrigrid method,
and we derive this system by minimising an energy functional of the form
\begin{equation}
\mathcal{E}=\int_{\Omega}\epsilon \frac{|\nabla u|^2}{2} + \frac{ u (1 - u)  }{\epsilon} + \Phi(u) + \lambda u \cdot F(c,I) ~dx,
\label{E}
\end{equation}
where
\begin{eqnarray}
\label{VV_fidelity}
F(c,I) = (I-c)^2~, ~~ c=\frac{\int_{\Omega}u I } {\int_{\Omega}u},
\end{eqnarray}
the quantity $c$ representing the average of $I$ in $u$, in other words being a measure of the oscillation of the data
over the support of $u$.

In order to achieve a simultaneous segmentation of $I$ into arbitrarily many pieces, we refer to the
vector-valued formulation of the Allen-Cahn system was introduced in Garcke, Nestler and Stoth
\cite{Garcke_Nestler_Stoth_97},
which allows one to consider an arbitrary number of components to the order parameter, now described
by a single vector-valued function $u \in \mathcal{V}^N$ in the function set defined as
\begin{equation}
\label{eqn_VN_space_defn}
\mathcal{V}^N \coloneqq \left\{ v \in \left( H^2(\Omega) \right)^N \colon \sum_{i=1}^{N}{v_i(x) = 1}\; a.e. \; in\; \Omega, \quad v_i \geq 0 \; a.e. \; in\; \Omega \right\},
\end{equation}
In other words, the vector-valued function $u$ must pointwise remain on the $N$-dimensional Gibbs Simplex
\begin{equation}
\label{Gibbs_simplex}
\mathbb{G}^N\colon=\left\{ x\in \mathbb{R}^{N} \left\arrowvert \sum_{i=1}^{N}x_{i}=1, \quad 0\leq x_{i}\right.\right\},
\end{equation}
which is itself an $(N-1)$-dimensional subset of the hyperplane
\[
\Sigma^N\colon=\left\{ x\in \mathbb{R}^{N} \left\arrowvert \sum_{i=1}^{N}x_{i}=1\right.\right\} .
\]
Trivially, $x_{i}\leq 1$ $\forall x \in \mathbb{G}^N$ (though not so for all $x \in \Sigma^N$).

An $N$-dimensional extension to the potential function $\Psi(u)$ is now required,
with $N$ minima given by $u_{j}=1, u_{i\neq j}=0$; the form
\[
Q(u)= \sum_{i=1}^{N}u_{i}(1-u_{i})
\]
has been used in the following, with $\Phi(u)$ now the indicator function on $\mathbb{G}^N$ and
$\Psi(u) = \Phi(u) + Q(u)$ as for the standard double-obstacle Allen-Cahn.

Following standard procedure, we wish to minimise the energy functional (\ref{E}) to derive a
pde to which we can introduce a pseudo-time stepping to find a global minimum.
To find the variational derivative of (\ref{E}) in a direction $v$,
care must be taken to remain in $\mathbb{G}^N$; in other words, we do not want our function to
leave the allowed set. If considering
$\mathcal{E}(u+\alpha v)$ for some $\alpha \in \mathbb{R}$, it must be that $\sum_i u+\alpha v = \sum_i u = 1$, and hence
$\sum_i \alpha v = 0$. To that end, Garcke, Nestler and Stoth \cite{Garcke_Nestler_Stoth_97} introduce the hyperplane
\[
T\Sigma^N\colon=\left\{ u\in \mathbb{R}^{N} \left\arrowvert \sum_{i=1}^{N}u_{i}=0\right.\right\}
\]
which is at all points tangent to $\Sigma^N$ (and hence to $\mathbb{G}^N$), together with the projection operator $T$ defined by
\begin{equation}
Tx \colon = x - \frac{1}{N}(x \cdot 1_N)\cdot 1_N
\label{projection_operator_T}
\end{equation}
as acting on a vector $x \in \mathbb{R}^N$, where $1_N \colon = (1, 1 \ldots 1) \in \mathbb{R}^N$. Geometrically,
the two hyperplanes $\Sigma^N$ and $T\Sigma^N$ are parallel; $T\Sigma^N$ passes through the origin;
the vector 
\[
\mbox{\^{N}}\colon =\frac{1_N}{N} ~ \in \mathbb{G}^N 
\]
is normal to $T\Sigma^N$ and represents the shortest distance between
$\Sigma^N$ and $T\Sigma^N$. As $N$ grows larger, this distance grows smaller. By construction,
$u-v \in T\Sigma^N$ $\forall u,v \in \Sigma^N$.

We now turn to the question of minimising (\ref{E}). As is usual for the Allen-Cahn functional (\ref{AC}),
we use a gradient descent method, i.e. we find the directional derivative of (\ref{E}),
and obtain a variational inequality to be solved numerically. Using the
notation $\langle \cdot, \cdot \rangle$ to indicate the standard inner product,
and using the the $N$-subgradient
\[
\partial \Phi (u) \colon = \left\{ \xi \in \mathbb{R}^N \; | \; \Phi(v) - \Phi(u) \geq \xi (v-u)  \right\},
\]
since we have
\begin{equation}
\label{eqn_xi_deduction}
\langle \xi,u-v \rangle \geq 0 \qquad \forall \, u, v \in dom \; \Phi,\; \forall \xi \in \partial \Phi(u).
\end{equation}
it is easy to see that a simple minimisation with respect to the
order parameter $u$, which subsequently determines the constants $c_i$ as well (as seen in \cite{Chan_Vese_98}), leads to
\begin{eqnarray*}
\langle \frac{\partial}{\partial u}\mathcal{E}(u) , Tv \rangle & \geq & \langle -\epsilon \triangle u -\frac{2}{\epsilon}u + \lambda F(c,I)+\frac{1}{\epsilon}\xi, Tv  \rangle \\
\end{eqnarray*}
and hence, by introducing a pseudo-time parametrisation, we have the inclusion
\begin{eqnarray}
\langle u_t -\epsilon \triangle u + T \left(-\frac{2}{\epsilon}u +\lambda F(c,I) \right), v-u \rangle & \ni & \langle - \partial\Phi(u) , v-u\rangle \\
& \geq & 0
\label{variation}
\end{eqnarray}
$\forall u,v \in \mathcal{V}^N$.
An approximation to this variational inequality can naturally be sought in terms of a finite element method, as described below.

\section{Discretisation and Numerical Solution}
\label{section_discretisation_and_numerics}

It has been shown that iterative solvers such as the Jacobi, Gauss-Seidel, Successive Over-Relaxation
and multigrid methods can be reduced to a class known as {\it Subspace Correction} methods; these first appeared in
Xu \cite{Xu_92}, see also Xu \cite{Xu_01}, Kornhuber \cite{Kornhuber_01}.
Convergence of subspace correction methods has been examined
for example in Tai and Xu \cite{Tai_Xu_01} for convex optimisation problems and in Neuss \cite{Neuss_98}.

In essence,
it is shown that a viable alternative to constantly projecting the solution from one level to another is to once
and for all project all basis functions from all multigrid levels onto the coarsest one, solving the problem by
iterating through all basis functions. An example of this projection method is shown in figure \ref{fig_basis_fn}.
This method was successfully applied to the vector-valued Allen-Cahn equation
by Kornhuber \cite{Kornhuber_03}, Kornhuber and Krause \cite{Kornhuber_Krause_03}, \cite{Kornhuber_Krause_06}.

In subsection \ref{finite_elements} we introduce the finite element method to establish the notation; building
upon that basis, subsection \ref{multigrid} shows the multigrid discretisation used, while subsection \ref{pseudocode}
details a few aspects of the numerical implementation due to the Gibbs Space constraint and the double-obstacle method.

\subsection{Finite Element Notation}
\label{finite_elements}

The continuous domain $\Omega$ is split into a set of subdomains $\mathcal{T}$, referred to as a
\textit{triangulation}, given by the set of triangles ${\tau}$ such that
\[
\overline{\Omega} = \bigcup_{\tau\in\mathcal{T}} \overline{\tau}
\]
The natural length scale associated with each triangulation is
\[
h \colon = \max_{\tau\in\mathcal{T}} \mbox{ } diam\mbox{}(\tau).
\]
The vertices of all triangles form a set of n points or {\it nodes}, and for the purposes of this work,
each triangulation is further required not to have any
\textit{hanging nodes}, i.e. nodes that are not corners of a triangle.
Each node $x_i \in \Omega, ~ i=1\dots ~n$ is assigned a function $\eta_i(x)$ such that
\[
\eta_i \colon = \left\{ \begin{array}{ll}
1 & \mbox{if $x=x_i$};\\
0 & \mbox{if $x\not= x_i$}.\end{array} \right.
\]
Although these functions can be arbitrarily elaborate while still satisfying the specified requirements,
continuous piecewise linear functions are more than adequate for second-order problems such as ours.
The set of all such functions, which can also be written as
\begin{equation}
\label{eqn_FE_nodal_basis_set}
\mathcal{S} \colon = \left\{ \eta \in C(\overline{\Omega})\colon \; \eta |_\tau \; \mbox{is linear} \; \forall \tau \in \mathcal{T}\right\} ,
\end{equation}
forms a basis for the finite element space
\begin{equation}
\label{eqn_FEspace}
\mathcal{V}^h=span \; \{\eta_i\}_{i=1}^{n}
\end{equation}
such that all functions $u^h \in \mathcal{V}^h$ can be represented as
\begin{equation}
\label{eqn_FEM_matrix_representation}
u^h(x)=\sum_{i=1}^{n}u^h_i \eta_i(x)
\end{equation}
and such that continuous functions $u$ on the original domain can be represented on the triangulation by
their piecewise linear interpolant $\pi \colon C(\overline\Omega) \rightarrow \mathcal{V}^h$.

We use the lumped mass and stiffness matrices
\[
\hat{M} \colon= \sum_{j=1}^{n}M = \langle \eta_i, 1 \rangle, ~ ~
A \colon = \langle \nabla \eta_i, \nabla \eta_j \rangle ~ ~
\forall \eta_i, \eta_j \in \mathcal{V}^h.
\]

To discretise the inequality in (\ref{variation}) by the finite element method, we pass to the the weak formulation
\begin{equation}
\label{eqn_weak_formulation}
\langle u^h_t +T \left( -\frac{2}{\epsilon} u^h + \lambda F(c,I) \right), v^h-u^h \rangle - \epsilon\langle \nabla u^h, \nabla (v^h-u^h) \rangle \geq 0 \qquad \forall v^h \in \mathcal{V}^h
\end{equation}
so that the discrete solution $u^h$ is only required to have $H^1$ regularity.
We can then consider the above problem as a series of sub-problems over each basis function.
This is known to be equivalent to an iterative solver, such as a Gauss-Seidel method;
the precise sequence in which these are considered can be altered for instance to follow a red-black Gauss-Seidel pattern.
By simple application of the properties of the projection operator $T$,
using the discretisation method outlined above, we have $n$ problems of the form
\begin{equation}
\label{E_fe}
\left( u_t + T \left( -\frac{2}{\epsilon} u + \lambda F(c,I) \right) \right) \hat{M} ( v - u) - \epsilon  u A (v-u)\geq 0.
\end{equation}
with some appropriate time discretisation to follow.
The inequality is due to the multi-valued nature of the subgradient at the boundaries of $\mathbb{G}^N$;
each iteration in the numerical method is performed as though (\ref{E_fe}) were a strict equality,
and if the result lies outside the acceptable space, then it is projected appropriately, as described in section \ref{pseudocode}.

\subsection{Multigrid Solution}
\label{multigrid}

\begin{figure}
\begin{center}
$\begin{array}{cc}
\multicolumn{1}{l}{\mbox{\bf (a)}} &
\multicolumn{1}{l}{\mbox{\bf (b)}} \\
\epsfxsize=2.2in
\epsffile{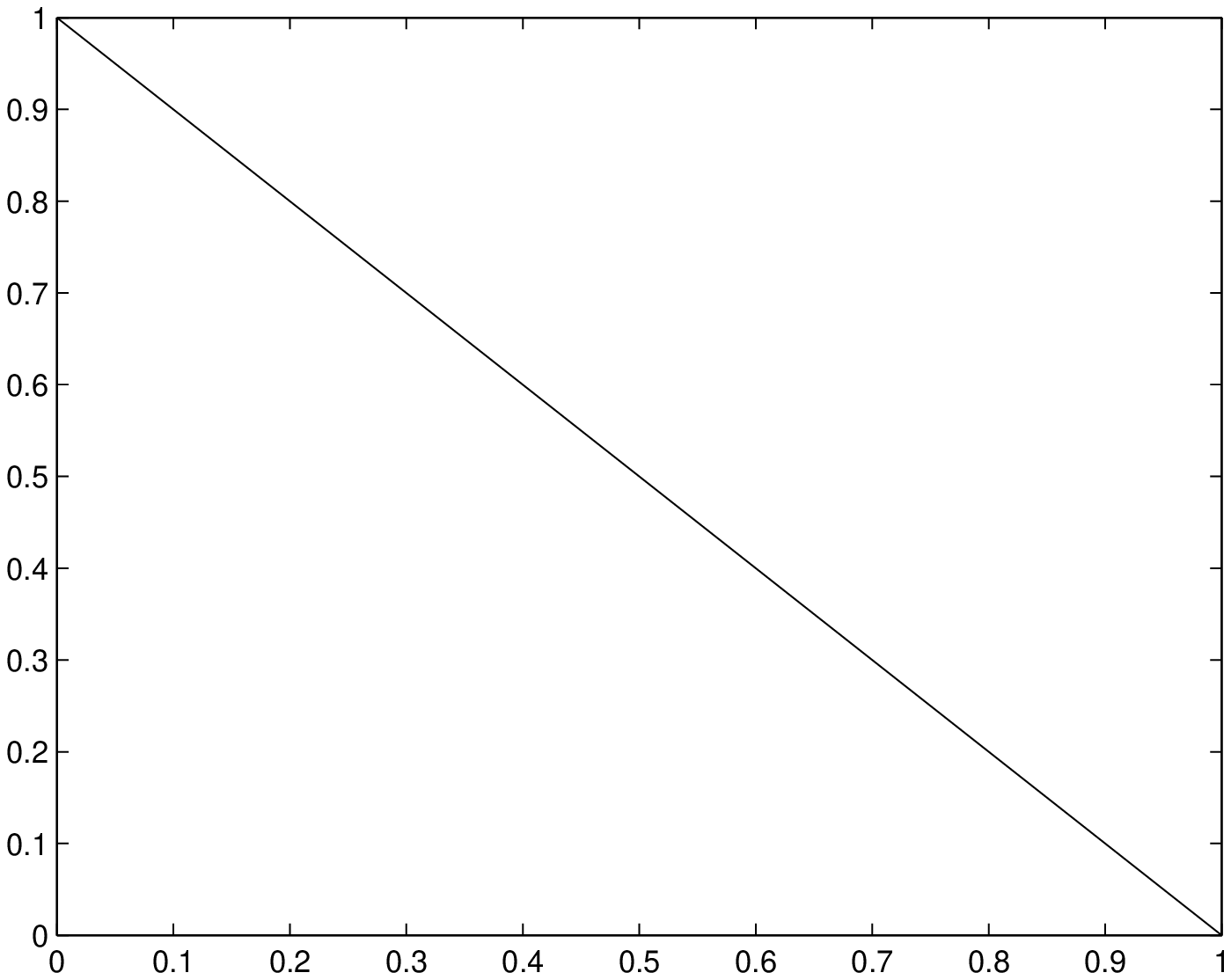} &
\epsfxsize=2.2in
\epsffile{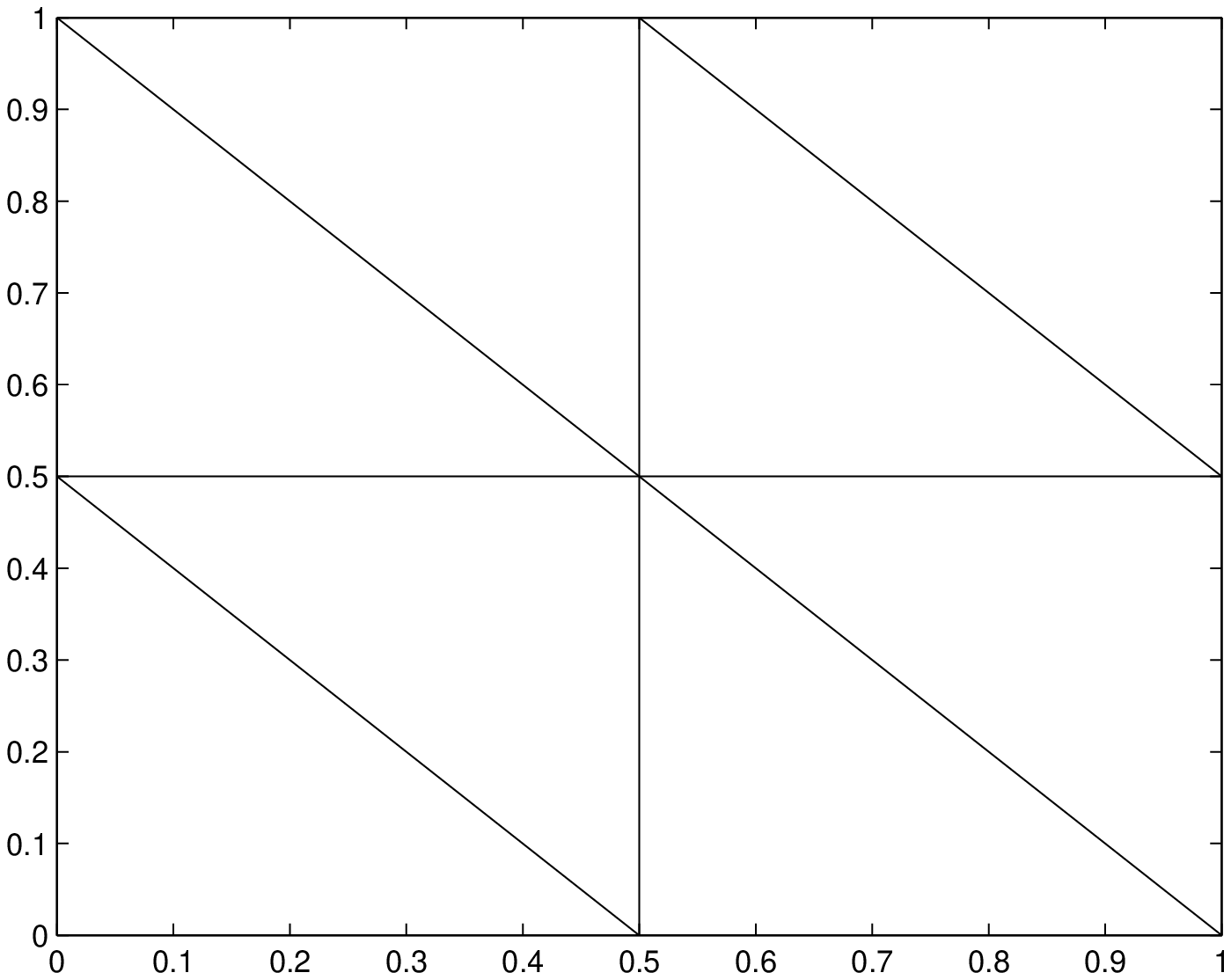} \\
\end{array}$
\end{center}
\caption{a) A coarse grid; b) the same grid, after one refinement.}
\label{fig_mesh_refinement}
\end{figure}

The rate of convergence of Gauss-Seidel methods requires $O(n)$ iterations per time-step,
and $O(n^2)$ computations overall to converge to a solution,
which is less than optimal.
Multigrid methods are known to be more efficient, in most cases being of only $O(n)$ complexity overall;
they rely on constructing several
nested finite element spaces, usually by refining a \textit{coarse} or \textit{macro} triangulation $\mathcal{T}_1$
$L-1$ times, eventually giving the \textit{fine} triangulation $\mathcal{T}_L$, where
\[
\mathcal{T}_l \subset \mathcal{T}_{l+1}.
\]
Figure \ref{fig_mesh_refinement} shows a simple example of one such refinement.
Each triangulation is associated with its own finite element function space. Using projection and restriction operators,
the solution is computed on one level, usually by a Gauss-Seidel solver or equivalent, and then passed to another level
to be corrected. The theoretical basis of the multigrid method lies in showing that the error at each iteration can be
considered to have several components of varying frequency, and that each time the problem is solved on a grid of
natural spacing $h$, the error components that are of frequency $h$ or higher are all reduced significantly, while those
with a lower frequency are barely affected. The multigrid method significantly improves convergence
by using several levels to deal with many components of the error in every iteration.

A recent development in multigrid methods has been to consider iterative solvers as part of the \emph{Succesive Subspace Correction}
framework, which simply considers a minimisation over a sequence of function spaces such as (\ref{eqn_FEspace}), each defined as the
set spanned by the basis functions defined on a different grid. In the context of finite elements, such a hierarchy of
function sets is readily provided by the basis functions at each level of refinement. To be more precise, given the problem
$\mathcal{P}_{i,j}$: find $u$ such that
\[
\langle u^h_t +T \left( -\frac{2}{\epsilon} u^h + \lambda F(c,I) \right), \eta_i-u^h \rangle - \epsilon\langle \nabla u^h, \nabla (\eta_i-u^h) \rangle \geq 0 \qquad \eta_i \in \mathcal{V}^{N,j},
\]
an $SSC$ method applied to this case consists of simply solving all such $\mathcal{P}_{i,j}$ by looping through all nodes $i$ and levels $j$
and updating the solution at each step, in whatever appropriate sequence the specific method demands.

Consider the finite element discretisation in (\ref{E_fe});
we firstly discretise in time using a backward Euler scheme to obtain the fully  
discrete problem.
At each time step, $j$, we use the SSC method to efficiently update the  
approximation $u^k$ as follows: starting at the coarsest level and moving  
to the finest in a standard multigrid \emph{w} pattern, for every basis function $\eta^i_L$, $ i=1,2, ...,  
n_L$ on level $L$ we update $u^{k+1} = u^k + \alpha^i_L$
where $\alpha^i_L$ satisfies the inequality
\[
\hat{M} \left( \frac{u^k + \alpha^i_L \eta - u^j}{\delta t} + T \left( -\frac{2}{\epsilon} u^j + \lambda F(c^j,I) \right) \right) - \epsilon A(u^k + \alpha \eta) \geq 0
\]
Once all basis functions and levels have been looped over, the iteration is complete; if the solution satisfies
some prescribed error tolerance, it is accepted and becomes the new iterate $u^{j+1}$; otherwise, the iteration
is repeated using the computed iteration as the new starting point.

\begin{figure}
\begin{center}
$\begin{array}{cc}
\multicolumn{1}{l}{\mbox{\bf (a)}} &
\multicolumn{1}{l}{\mbox{\bf (b)}} \\
\epsfxsize=2.2in
\epsffile{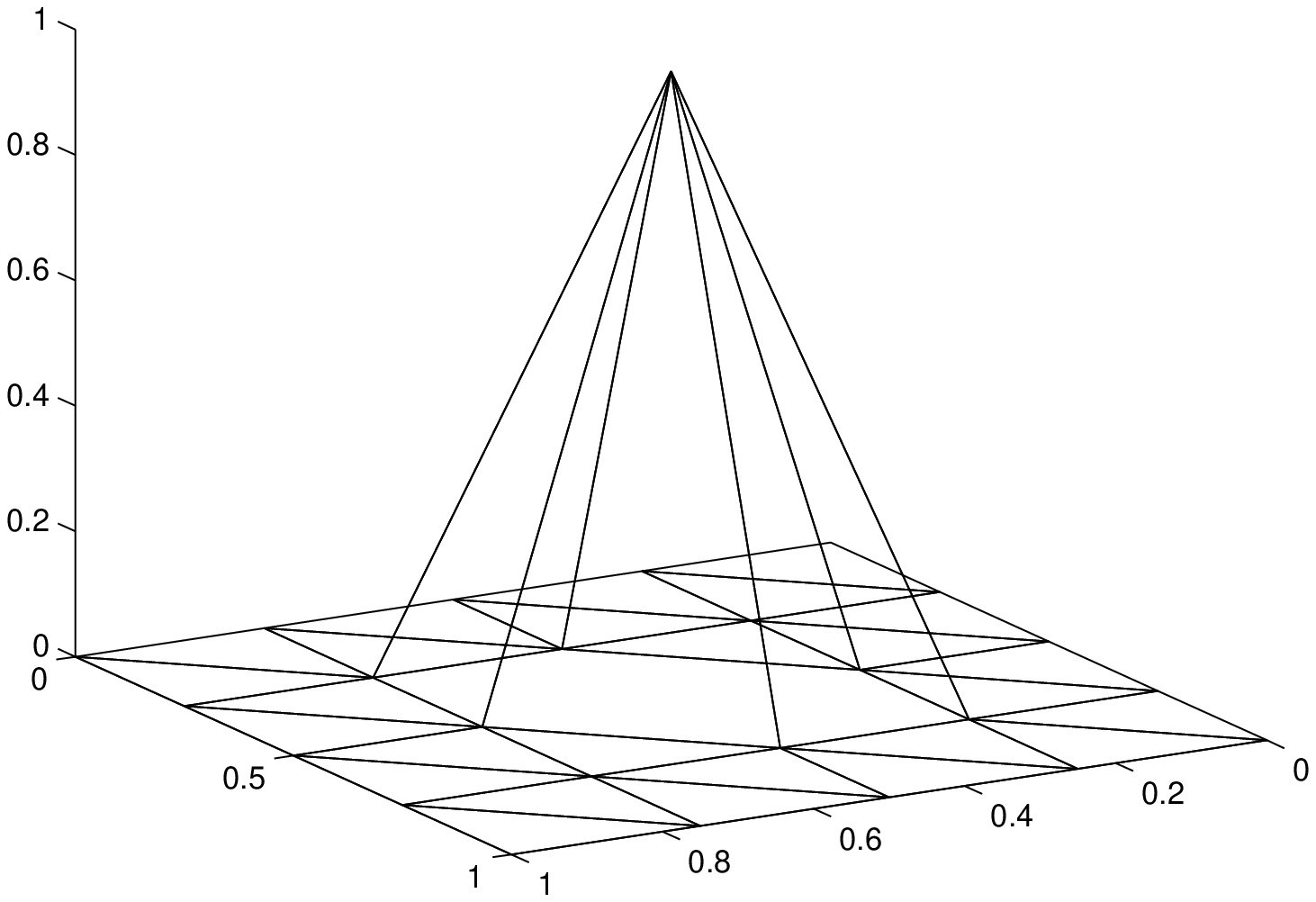} &
\epsfxsize=2.2in
\epsffile{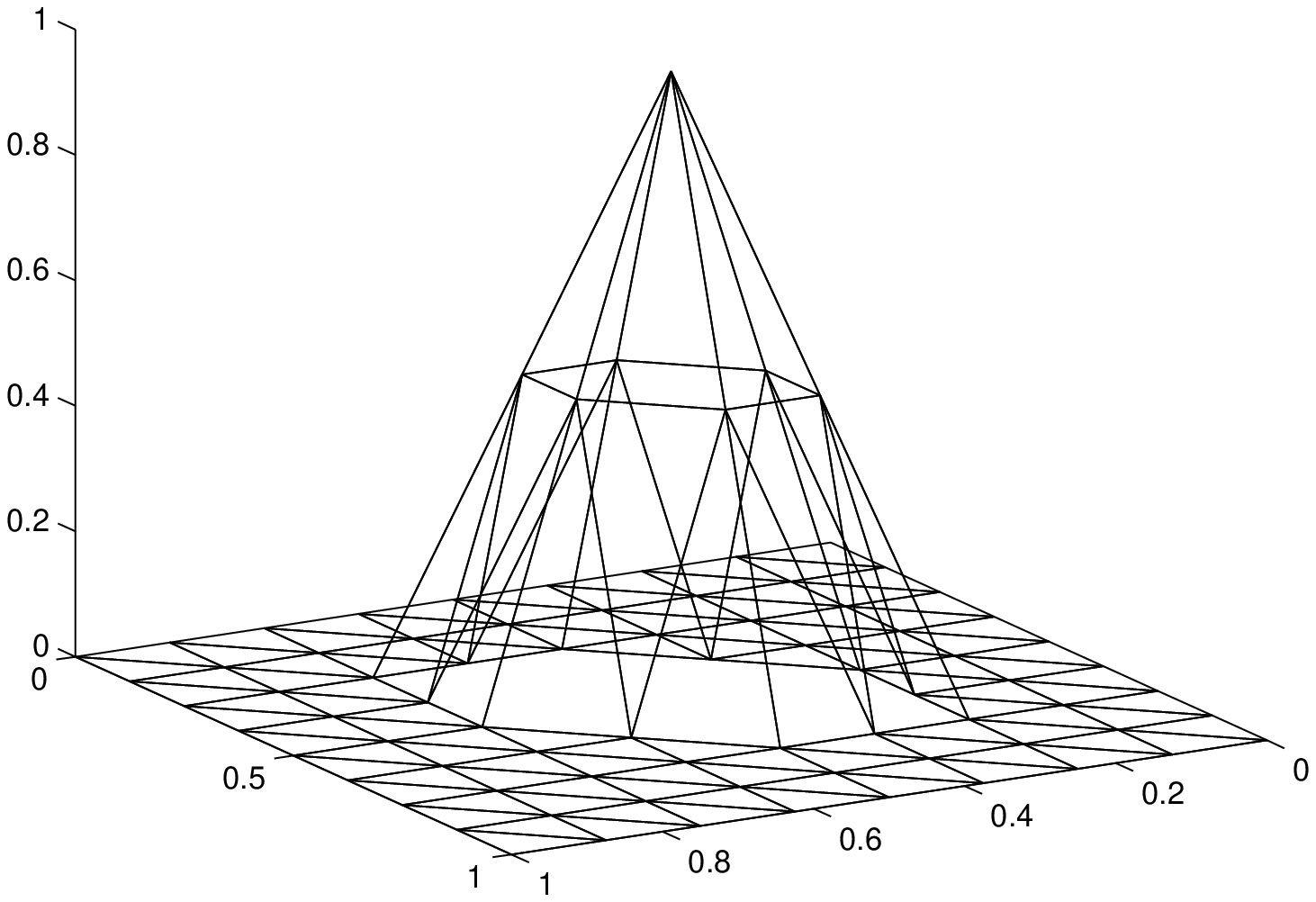} \\
\end{array}$
\end{center}
\caption{a) A basis function defined on a coarse grid; b) the same basis function projected onto a finer grid}
\label{fig_basis_fn}
\end{figure}

All basis functions $\eta$
of all multigrid levels are projected onto the finest grid. An example of this projection is given in figure \ref{fig_basis_fn}.
Thus each iteration is of the form
\[
\hat{M} \left( \frac{u^k + \alpha \eta - u^t}{\delta t} + T \left( -\frac{2}{\epsilon} u^t + \lambda F(c^t,I) \right) \right) - \epsilon A(u^k + \alpha \eta) = 0
\]
before projection into the required function space.
It is enough to multiply both sides by $\eta^T$ and re-arrange the above by simple algebra to obtain
an equation for $\alpha$:
\begin{equation}
\label{alpha}
\alpha \eta^T ( \hat{M} - \epsilon \delta t A  ) \eta  = \eta^T \left\{  \hat{M} \left[ u^t - u^k + \delta t T \left( \frac{2}{\epsilon}u^t - \lambda F(c^t,I) \right) \right] + \epsilon A u^k \right\} .
\end{equation}

\subsection{Gibbs Space Constraint Pseudocode}
\label{pseudocode}

In the vector-valued case, since $N$ functions need to be updated simultaneously for each basis function,
it is clear that $\alpha$ is also vector-valued and of size $N$. In order to preserve the constraint
\[
\sum_{i=1}^{N}u_i = 1.
\]
it must be that for each update
\[
\sum_{i=1}^{N}\alpha_i = 0.
\]
This is actually taken care of by the choice of minimisation, i.e. by involving the projection operator $T$ and
only allowing search directions $u + \alpha T v$.

Secondly, for the functions to remain in the Gibbs space $\mathbb{G}^N$ at all times, at each iteration of the numerical
scheme we must ensure that the result still lies in $\mathbb{G}^N$, i.e. in addition to the restriction on the
sum of $\alpha$ above, all components of $u$ must be positive. This requirement may be less than straightforward
to enforce on coarse grids, where several nodes are affected by the change in $\alpha$; this is also mentioned in
Kornhuber \cite{Kornhuber_03} and is due to the large support of the coarse grid basis functions
after projection onto the finest grid; for that reason, a truncated multigrid algorithm may be more efficient than
a full v-cycle.
 
The pseudo-code for a general subspace correction is given below;
it is assumed that for each basis function, a list of affected nodes has been drawn so that the least
amount of work possible is being done.

\begin{enumerate}

\item Evaluate $\alpha$ as per (\ref{alpha}); then, calculate a trial version of $u$.

\item Check for each function $u_i$ whether any of its entries have become negative.
 If not, no further work is required.

\item Otherwise, we have a split of the functions $u_i$ into two sets, $\mathbb{Q}$ being the set of $u_i$ with negative 
(unacceptable) values, and $\mathbb{P}$ being the remainder.

For each function in $\mathbb{Q}$, we need to determine a new $\alpha_i$ 
such that $u_i$ remains between 0 and 1, i.e. such that the lowest value is zero. This is easily done and we refer
to this correction as $\beta$.
Since we know that $\sum \alpha = 0$ must be satisfied, if we decrease $\alpha_i$ then we need to increase all other 
$\alpha_j, j\neq i$ by a corresponding amount. However, we can exclude those functions that are already negative.
Therefore, if there are $k$ functions in $\mathbb{P}$ and we decrease $\alpha_i$ by an amount $\beta$,
then we need to decrease all $\alpha_k$ by $\beta / k$.

Do this for each function in $\mathbb{Q}$.

\item All functions in $\mathbb{Q}$ are now guaranteed to lie between 0 and 1.
However, the successive corrections $\beta$ applied to the functions in $\mathbb{P}$
may have caused some of those to become negative; hence, repeat from step 2, but only considering
a restricted set of functions. Repeat this process until there are no more negative functions.

\end{enumerate}

Since at least one function is being removed from the set of all available functions at each repetition of this process,
it is a procedure of maximum order $N$ complexity.

\section{Implementation}
\label{section_implementation}

\subsection{Determination of Natural Scales}

In principle, it is necessary to consider the possibility that the interface between two areas of interest be
only one pixel wide; in other words, that two neighbouring pixels belong to different objects in the image.
It is well known that the interface width of the double obstacle Allen-Cahn equation is of $\mathcal{O}(\epsilon)$ -
see for example \cite{Elliott_97} and references therein. This immediately suggests a natural length scale for the
problem: an $\epsilon$ smaller than the size of a pixel in the given data would not make sense, and a larger $\epsilon$
would only blur edges and corners. All our computations have been performed using this natural length scale.

The value of $\epsilon$ also suggests an estimate for a plausible $\lambda$; if the value $\sigma = \lambda \epsilon$ were
much less than $1$, the effects of the fidelity terms would be negligible; conversely, if it were too much larger than $1$,
the interface width would be reduced to such a slim margin that it could no longer be considered a diffuse interface.
It was found in our computations that a value of roughly $\sigma \in [10, 100]$ gave the most interesting results.

However, it must be noted that the numerical discretisation of the
Allen-Cahn equation can only give interface motion results of a satisfactory level of accuracy when there are several
nodes to represent each interface, ideally $8$ but in practice at least $4$. In the context of a multigrid simulation, this can be 
achieved quite naturally by taking the given image's pixel structure as the reference grid and refining it an appropriate
number of times - assuming uniform refinement to be the simplest case, where the number of nodes roughly doubles
each time, this implies at least three refinements. It is also possible to coarsen the regular pixel grid a few
times, depending on the size of the image. Thus the coarsest grid in the final setup will probably
not correspond to the grid on which the data was defined.

Another consequence of the need for a refinement process is the need to represent the image data on a finer grid
than the one it was defined on. In the present model, the only link between the evolution of the function $u$ to the
image to be processed are the fidelity terms of the form
\[
\int_{\Omega} (I - c)^2 ~dx
\]
with the constants
\[
c=\frac{\int_{\Omega}I u} {\int_{\Omega}u}
\]
approximating the average of $I$ on $supp \, u$.
The method of implementation of these terms is therefore critical to obtaining a good result. Moreover,
a decision needs to be made as to what function space the image $I$ is assumed to belong to. There
are several candidates, such as $C^\infty (\Omega)$ smooth functions, which make the mathematical
analysis much easier and are generally derived by convolving the image data with a Gaussian;
Lipschitz continuous functions, which include piecewise linear interpolants; and functions of bounded
variation. This distinction can be of great practical importance to a finite element method
because depending on the chosen approach to node placement and fidelity term implementation
it may be necessary to compare the values of $u$ and $I$
at points that do not correspond to given pixel values, or to points that lie exactly on the boundary
between two or more pixels (e.g. corners). Which space the image belongs to ultimately depends on what kind
of data one is examining, but in this case we have chosen to consider the image as a set of piecewise constant
values over each pixel; this is done in order not to artificially blur any edges before any computations
have even taken place.

Together with mesh refinement, it is necessary to project the values of the original pixels to the fine grid. This is
not done by interpolating the data values in any way, in order not to create spurious gradients. This is
necessary because the fitting terms in our method rely on calculating the average value of $I$ in each region;
introducing new gradients also introduces small areas of different average value, which may be
erroneously identified as new objects. Therefore, every newly created node is assigned
exactly the same value as one of its neighbours. This preserves sharp discontinuities; it also leads to some staircasing
of the data, but only on a relatively small scale that should be locked out by fixing $\epsilon$ on the coarse grid.

\subsection{Data Projection}
\label{interpolation_and_post-processing}

If it is indeed necessary to project the values of the original pixels to the fine grid, this should
not be by interpolating the data values in any way, in order not to create spurious gradients. This is
necessary because the fitting terms rely on calculating the average value of $I$ in each region;
introducing new gradients also introduces small areas of different average value, which may be
erroneously identified as new objects in their own right. Therefore, every newly created node is assigned
exactly the same value as one of its neighbours, which preserves sharp discontinuities. In other words,
the image data is taken to represent an underlying function $I \in BV$ of bounded variation.

There are at least two distinct options for the implementation of these terms using a finite element model.
The first is direct projection of the image data, node by node. The second is to note that the image function
$I$ is only weakly represented in the FE system, in the sense that only the value of its integral over a region
is contained in the fidelity terms, and therefore in principle only the value of its integral on each element
is required; projection can therefore be carried out triangle by triangle. This makes perfect sense as long as the
triangulation is nested. In other words, the fidelity terms can be
computed by either of the two following methods: either one uses the standard mass matrix,
\begin{eqnarray*}
M_{ij}\colon=\langle \eta_i, \eta_j \rangle,\\
F = \sum_{n=1}^{N} M_{in} \cdot \left( I_n - \frac{M\cdot I \cdot u}{\sum M \cdot u}\right)^2
\end{eqnarray*}
or one computes the integral of $I$ from first principles:
\begin{eqnarray*}
MI = \sum_{n=1}^{N} \sum_{t=1}^{\tau_n} I|_t \langle \eta_i|_t, \eta_n|_t \rangle; \\
F = \left( MI - \frac{MI\cdot u}{\sum M\cdot u} \right)^2
\end{eqnarray*}

\begin{figure}
\begin{center}
\mbox{Projection by node}\\
\epsfxsize=4.2in
\epsffile{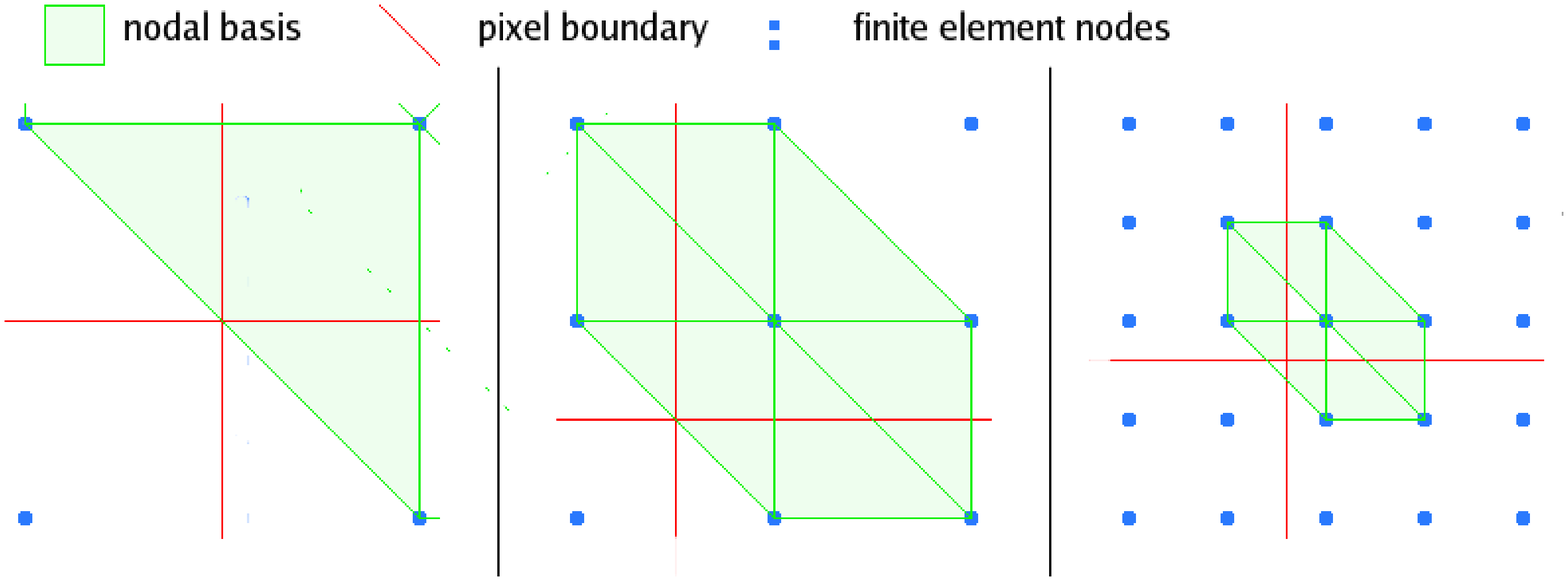}\\
\mbox{Projection by simplex}\\
\epsfxsize=3in
\epsffile{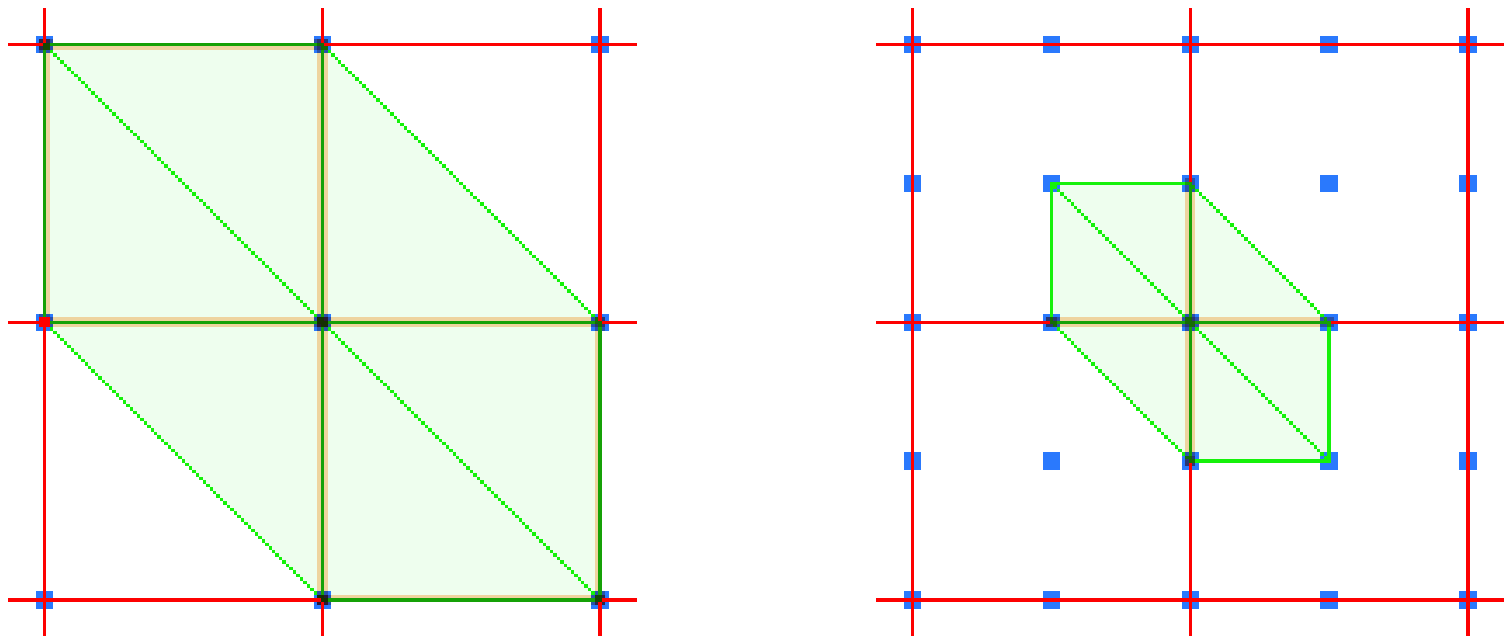}\\
\end{center}
\caption{A clarification of two different data projection schemes}
\label{fig_projection}
\end{figure}

Each method is associated with its own advantages, disadvantages, and computational costs. It is worth
noting that the errors associated with each one decrease with each mesh refinement.
The former can be thought of as \emph{projection by node} and the latter as \emph{projection by simplex}; examples are
shown in figure \ref{fig_projection}.

%
%

%
%

\subsection{Post-processing}

\begin{figure}
\begin{center}
$\begin{array}{cc}
\multicolumn{1}{l}{\mbox{\bf (a)}} &
\multicolumn{1}{l}{\mbox{\bf (b)}} \\
\epsfxsize=1.8in
\epsffile{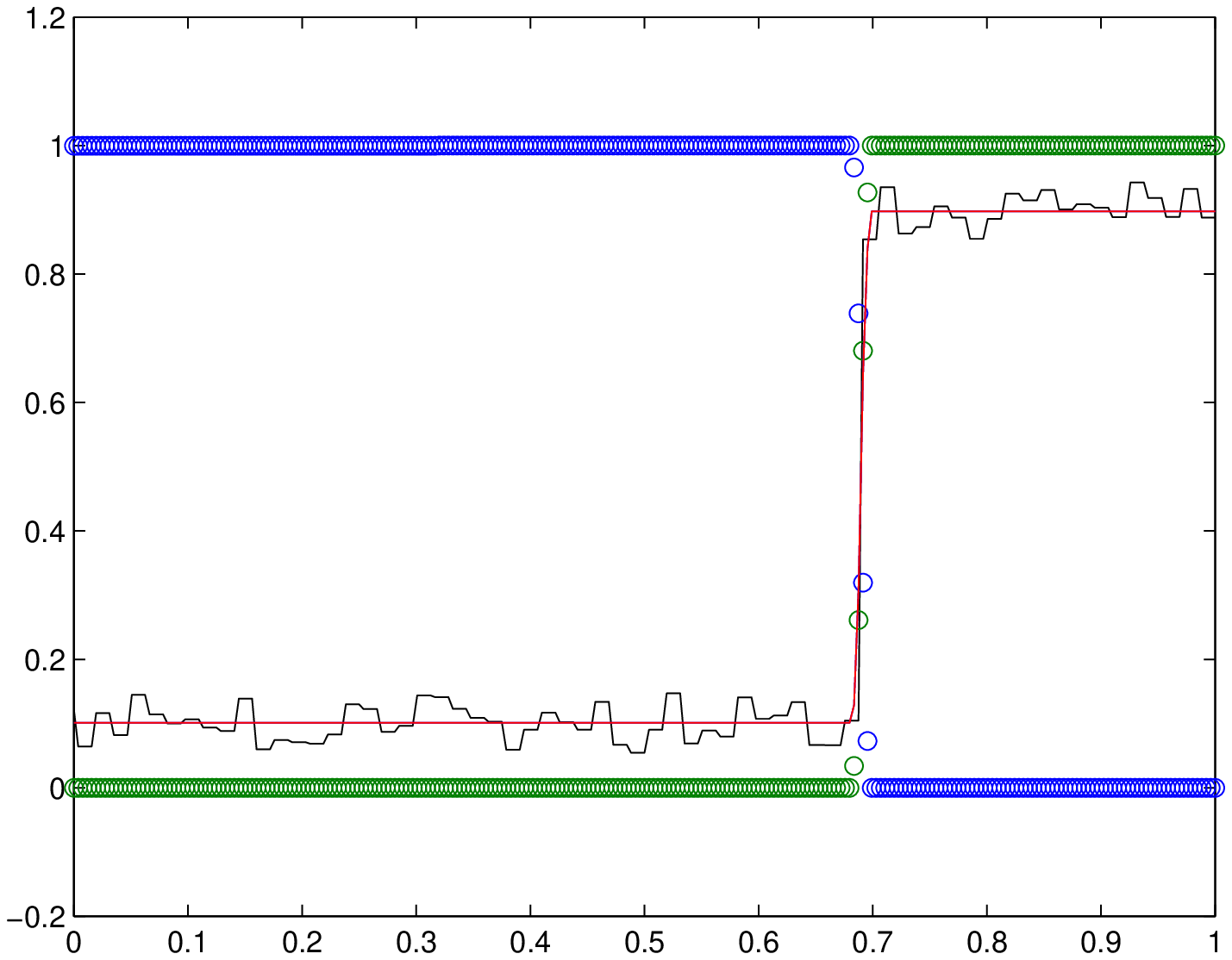} &
\epsfxsize=1.8in
\epsffile{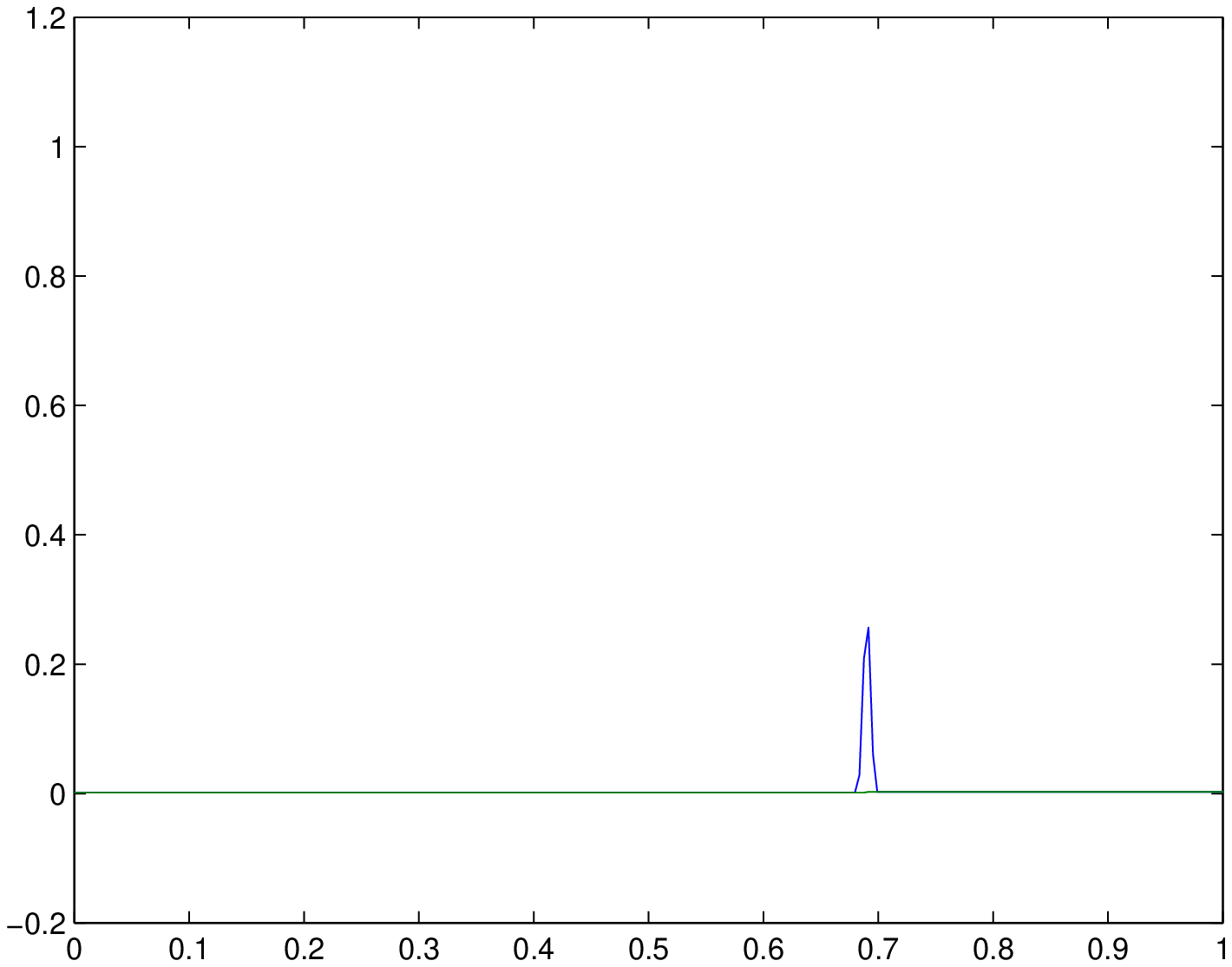} \\
\end{array}$
\end{center}
\caption{a)A comparison of noisy data (black), segmented regions (green and blue), and resulting denoised signal (red); b) errors for the composite (blue) and rounded composite (green) segmentation.}
\label{projection_01_02}
\end{figure}

Fig. \ref{projection_01_02}\textbf{a} shows an example of the post-processed solution.
The green and blue functions indicate segmented
regions as identified by the program; using these, the average of the data, $osc$ $I$, is used
in each to obtain the denoised data (red). This process is easily achieved by multiplying said average by the component
itself, then summing all the resulting components; this is referred to as the \textit{composite}. 
Recall that the measure of $osc$ $I$ on the support of each component $u_i$ is already known and used
actively in the fitting terms driving the evolution (see (\ref{VV_fidelity})). Further, because
each component has values not identical to $0$ or $1$, notably at each interface, it is useful to round
all values to either extremum, in such a way that only one component is equal to $1$ and all others are $0$ at any
given point; in this way, segmented regions are defined more precisely.
This naturally leads to the \textit{rounded composite}, the advantage of which is shown
in \ref{projection_01_02}\textbf{b}, a comparison of the errors given by the composite and rounded composite.

\begin{figure}
\begin{center}
$\begin{array}{cc}
\multicolumn{1}{l}{\mbox{\bf (a)}} &
\multicolumn{1}{l}{\mbox{\bf (b)}} \\
\epsfxsize=1.8in
\epsffile{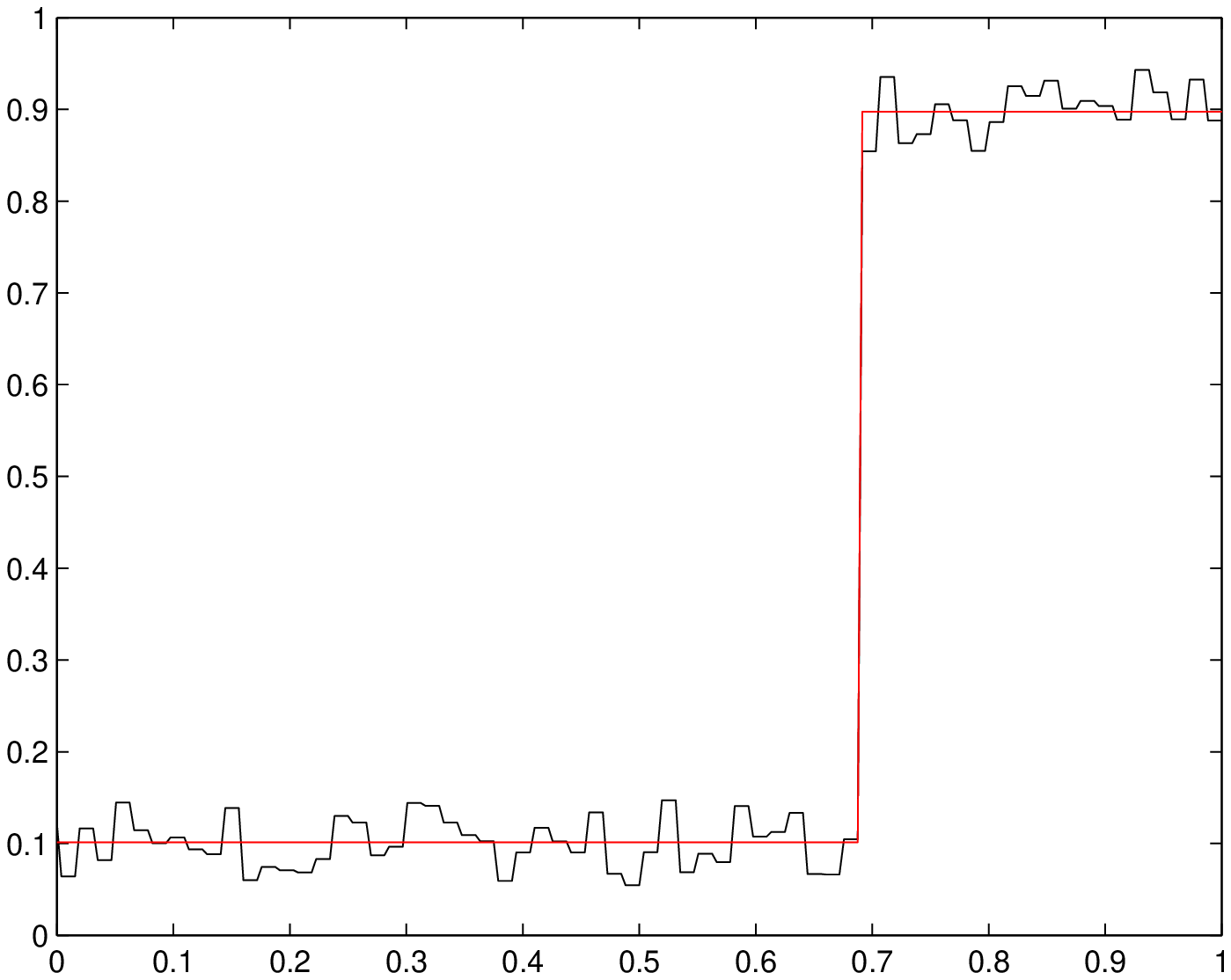} &
\epsfxsize=1.8in
\epsffile{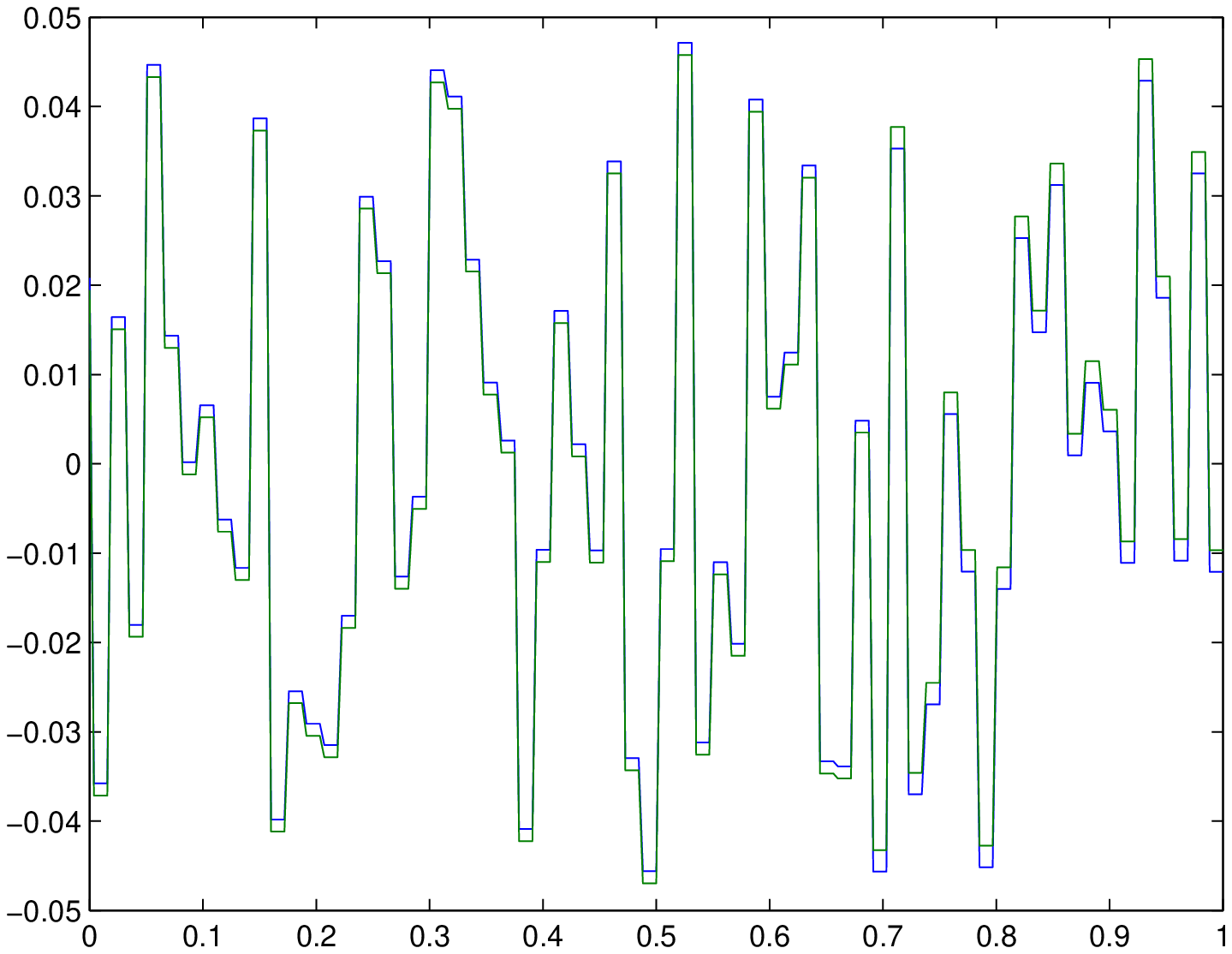} \\
\end{array}$
\end{center}
\caption{a) noisy data (black) and rounded composite (red); b)original noise (blue) vs. recovered noise (green).}
\label{projection_01_03}
\end{figure}

Figure \ref{projection_01_03}\textbf{a} shows the noisy data (black) and the resulting rounded composite (red);
for comparison, Fig. \ref{projection_01_03}\textbf{b} shows the original noise (blue) and the recovered noise (green) obtained
by subtracting the rounded composite from the initial data.

\subsection{Two Dimensional Examples}

\begin{figure}
\begin{center}
$\begin{array}{cc}
\multicolumn{1}{l}{\mbox{Image data}} &
\multicolumn{1}{l}{\mbox{Rounded composite}} \\
\epsfxsize=1.8in
\epsffile{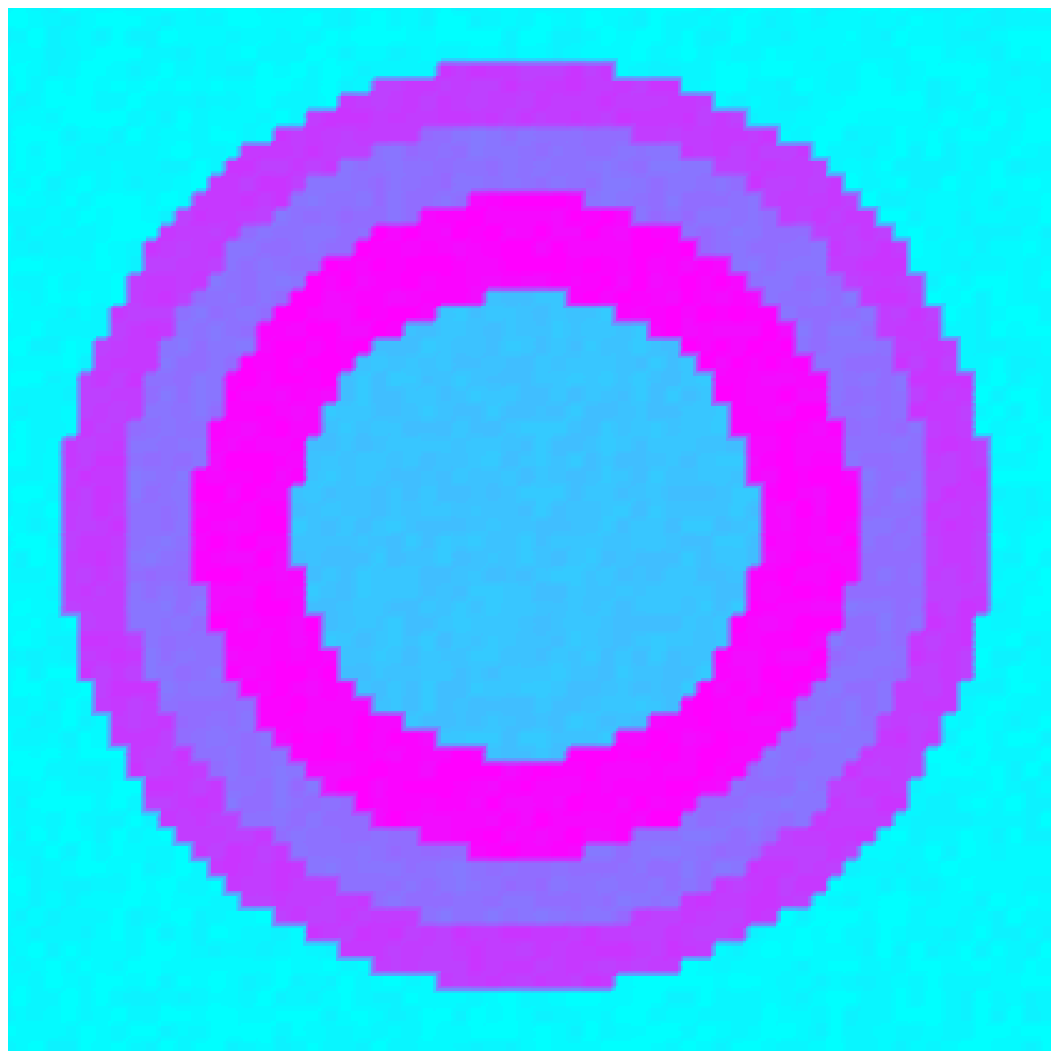} &
\epsfxsize=1.8in
\epsffile{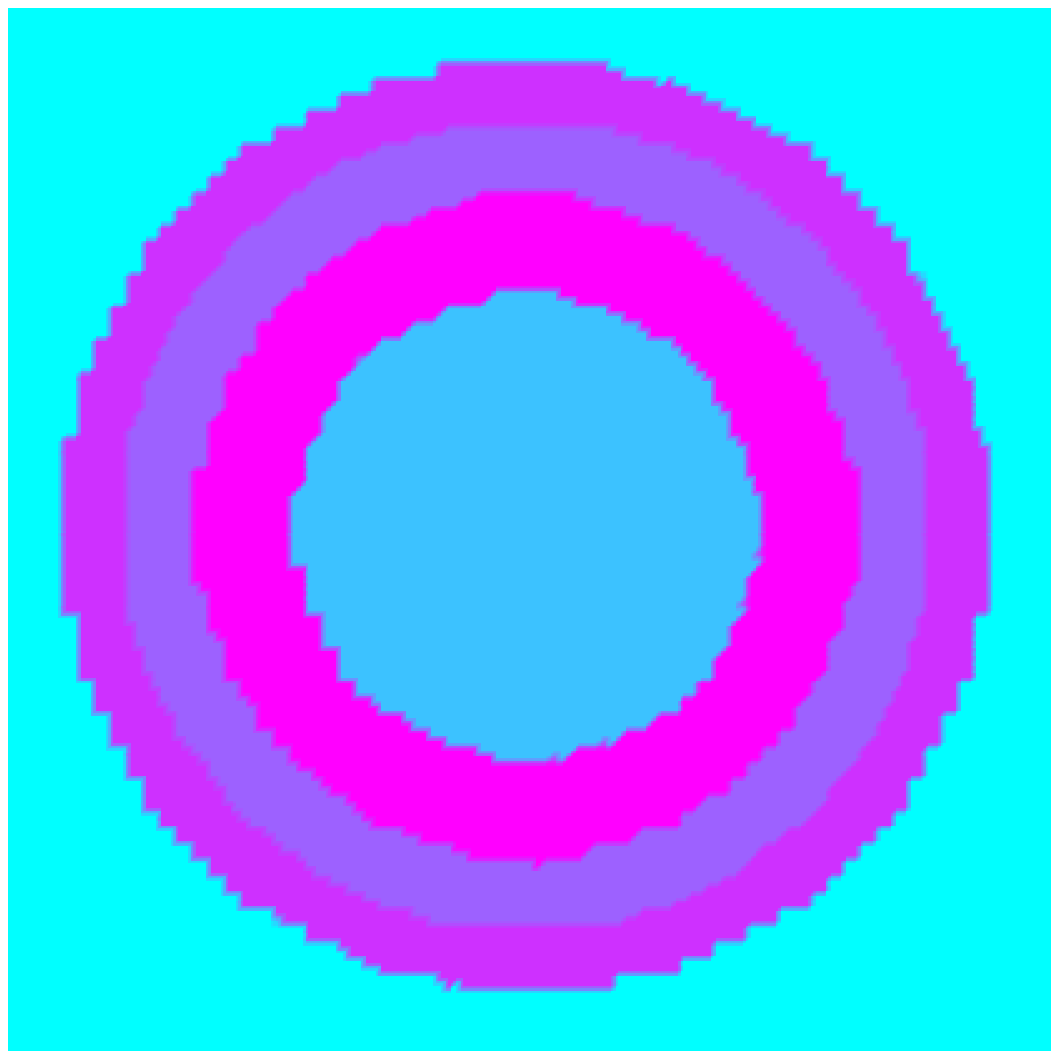} \\
\multicolumn{1}{l}{\mbox{Vectorial constituents}} &
\multicolumn{1}{l}{\mbox{Remainder}} \\
\epsfxsize=1.8in
\epsffile{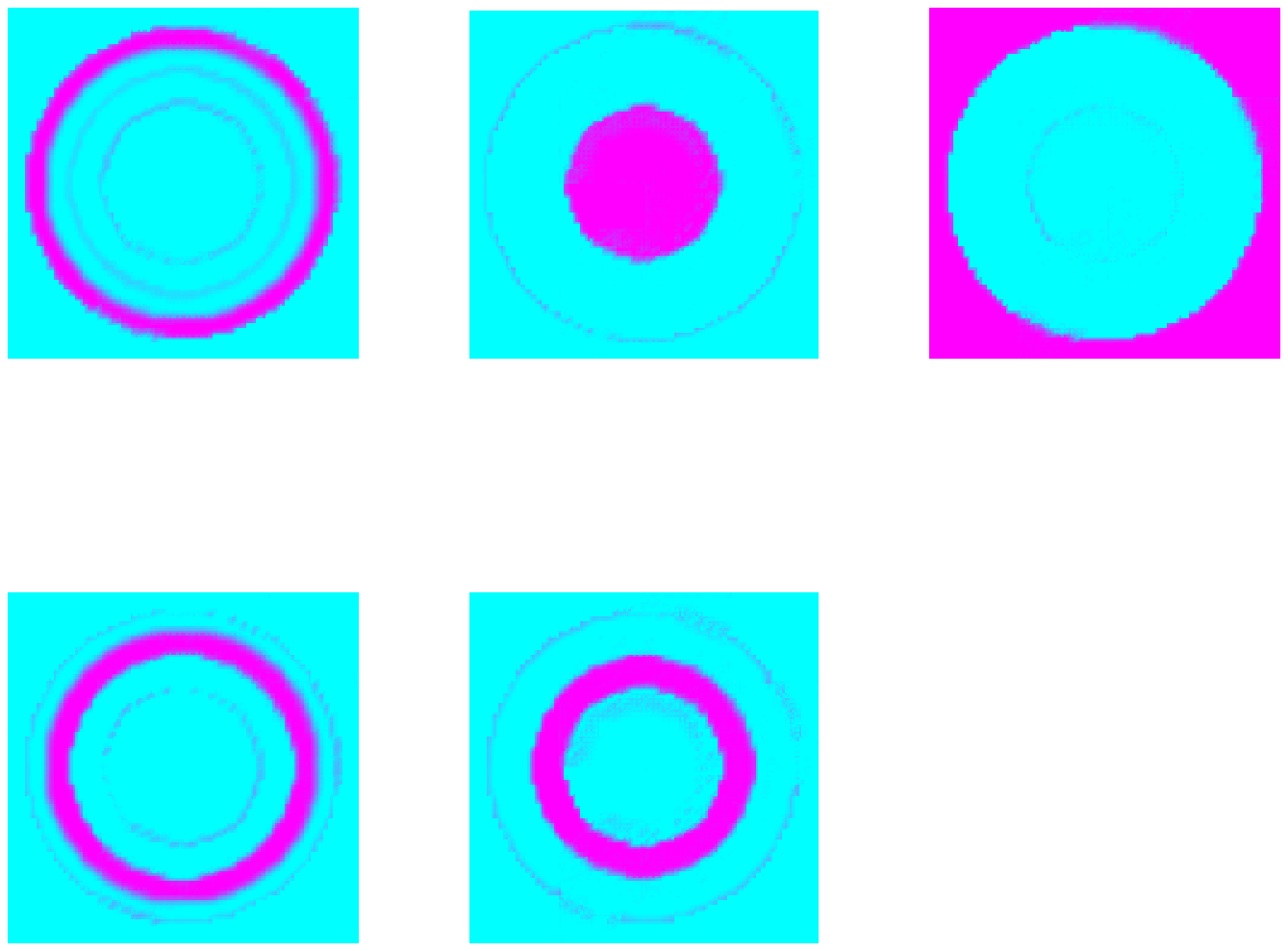} &
\epsfxsize=1.8in
\epsffile{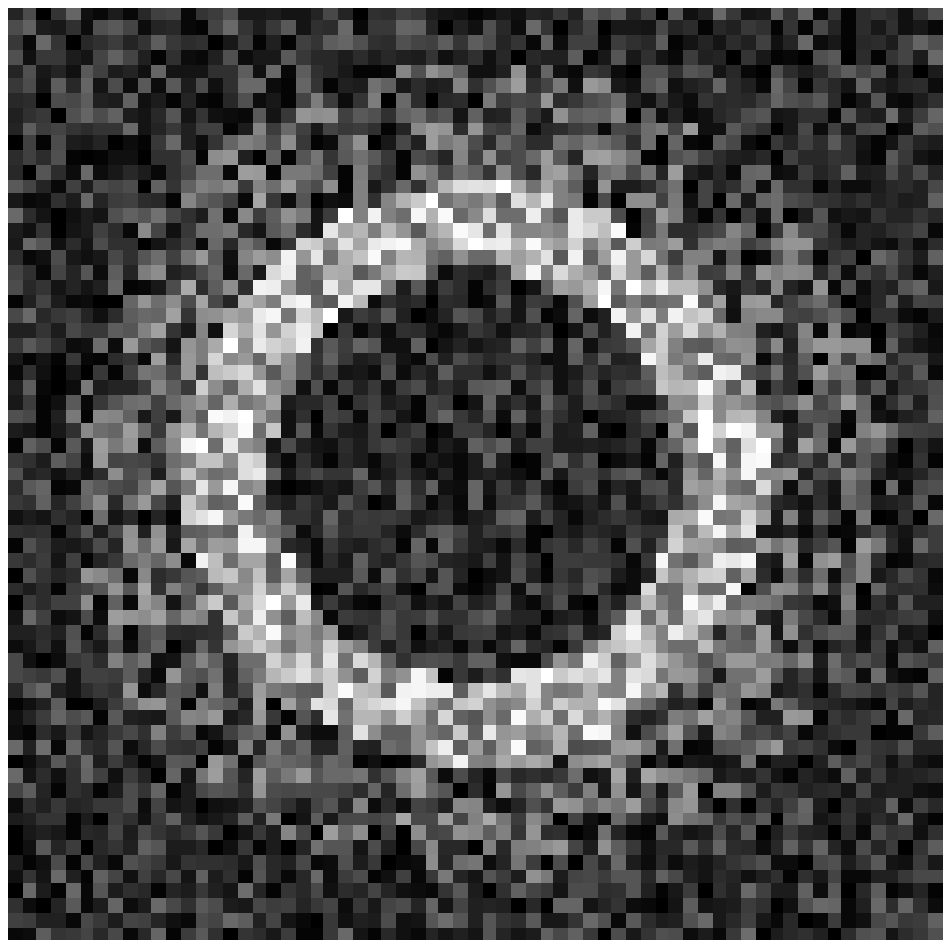} \\
\end{array}$
\end{center}
\caption{Segmentation of four concentric circles at [.25  .95  .55  .75] over a background at level .1, with added random noise
of amplitude .05. The remainder shows a combination of error and noise, scaled to show all detail} 
\label{circles01_01}
\end{figure}

\begin{figure}
\begin{center}
$\begin{array}{cc}
\multicolumn{1}{l}{\mbox{Image data}} &
\multicolumn{1}{l}{\mbox{Rounded composite}} \\
\epsfxsize=2.2in
\epsffile{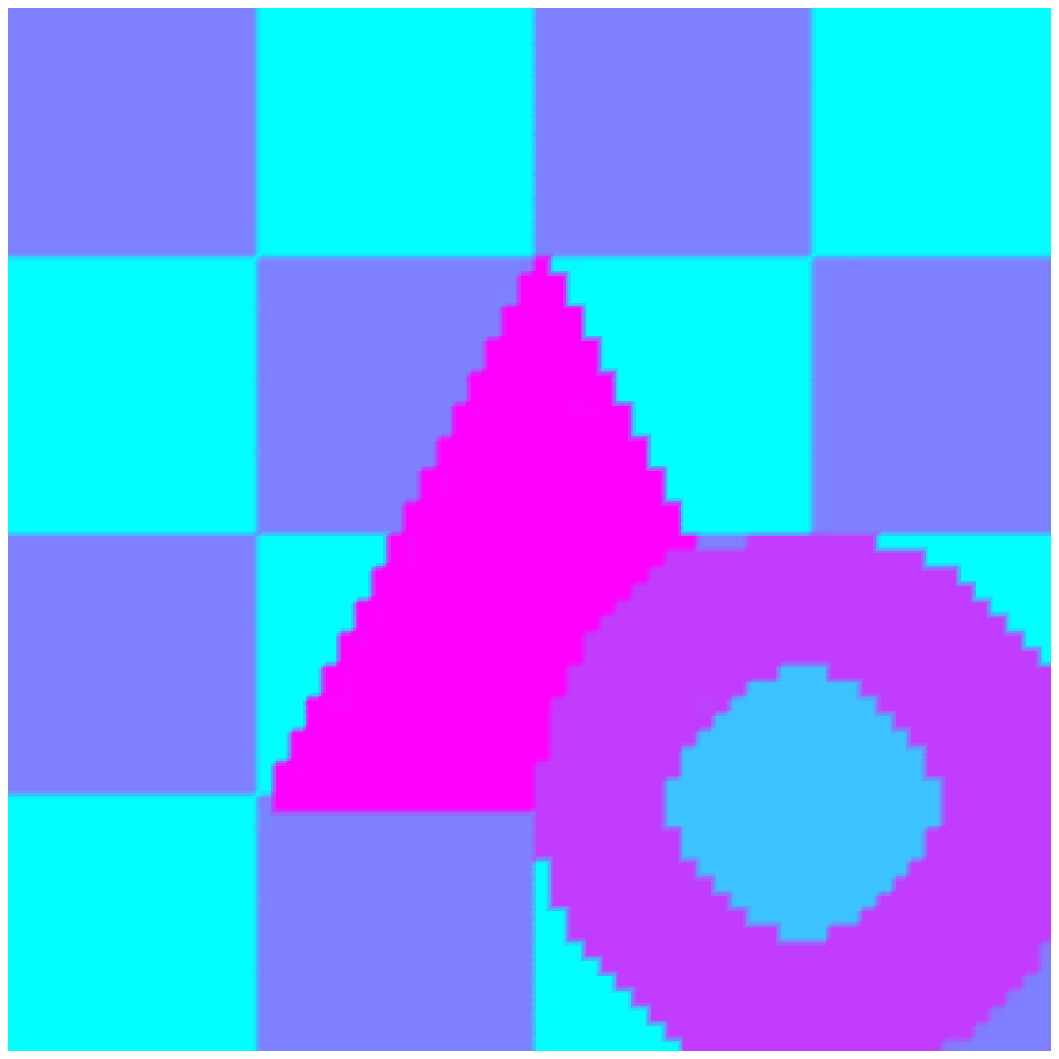} &
\epsfxsize=2.2in
\epsffile{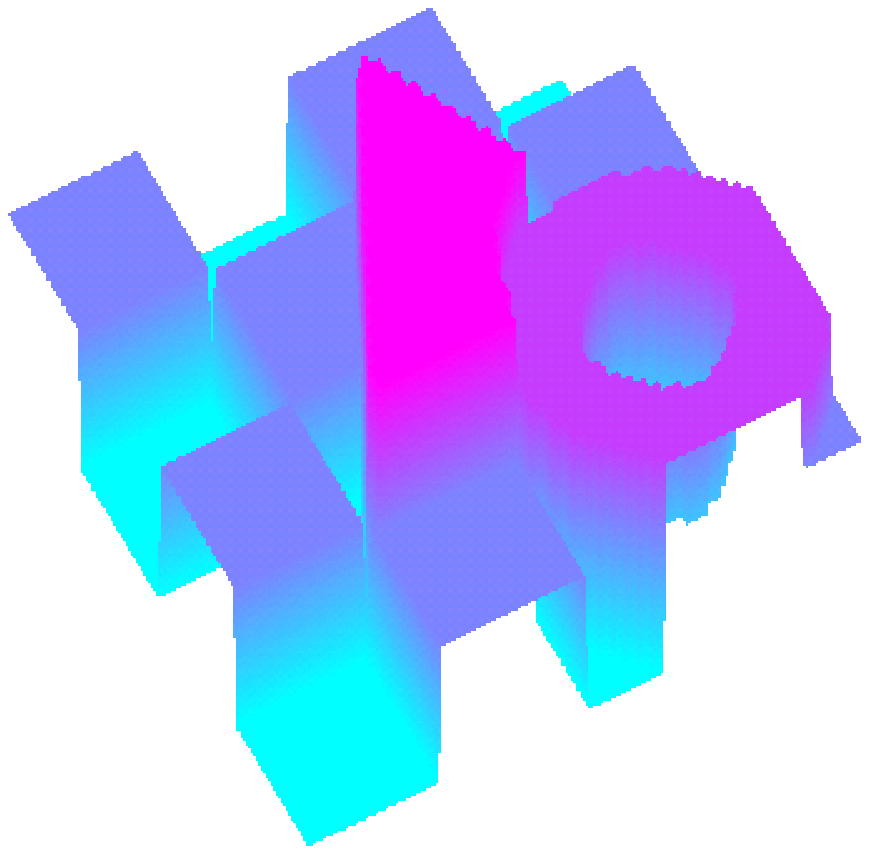} \\
\multicolumn{2}{l}{\mbox{Vectorial constituents}} \\
\multicolumn{2}{l}{
\epsfxsize=4.4in
\epsffile{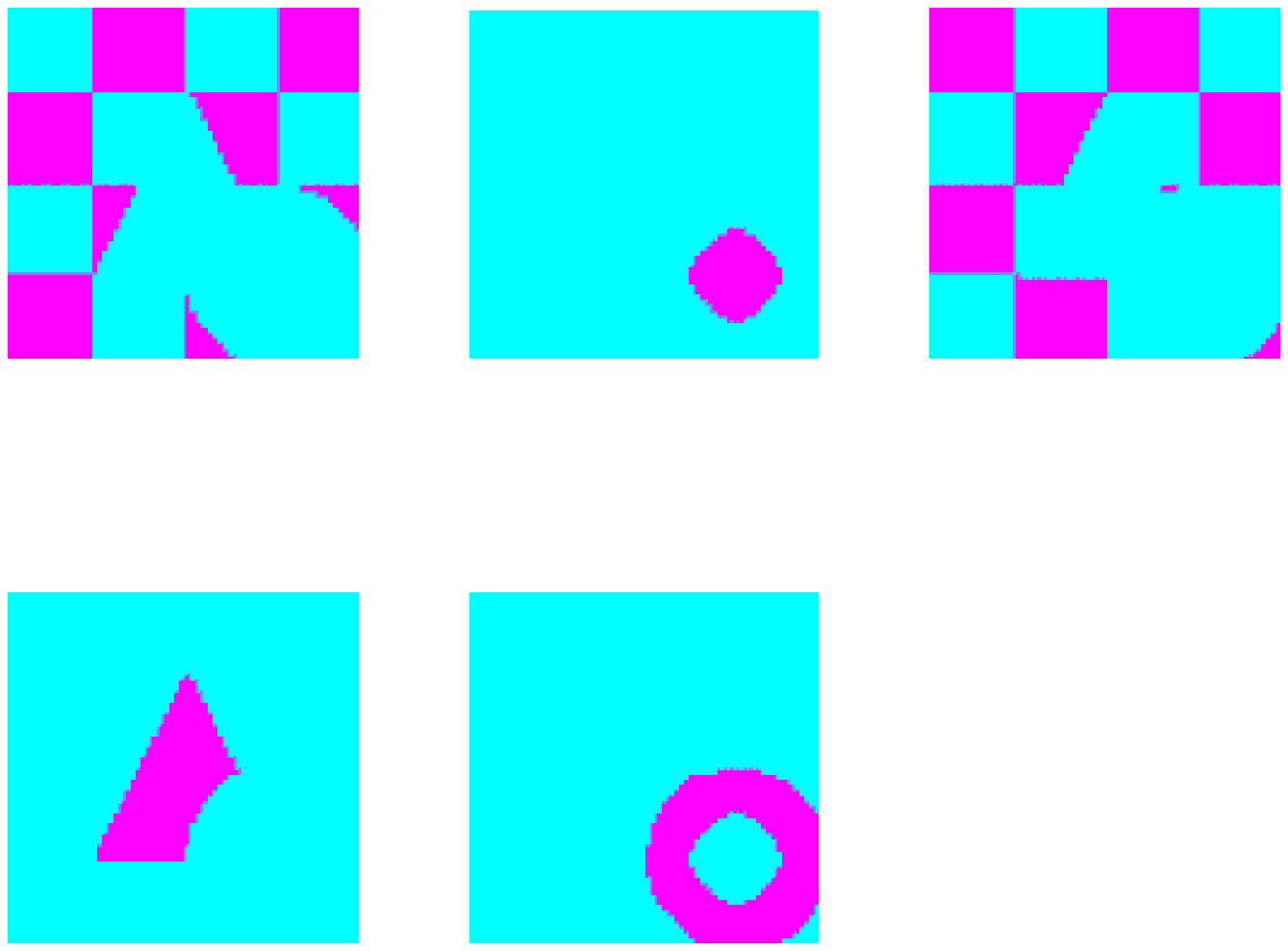} }\\
\end{array}$
\end{center}
\caption{Segmentation of a geometric composite} 
\label{chequerboard_65}
\end{figure}

Figure \ref{circles01_01} shows the results of our method as applied to an image of concentric circles, with some noise.
Figure \ref{chequerboard_65} shows a more interesting case depicting what appears to be several overlapping geometric solids
superimposed on a chequered background, for comparison with the results in Jung, Kang and Shen \cite{Jung_Kang_Shen_06}.

We also wish to examine the case of multi-channel data, the most obvious example of which
is that of colour images. Although there is no reason
why the proposed method could not be extended to images acquired at different wavelengths, for example by combining
a visible night-time photograph with an infrared scan of the same area, the following
arguments will concentrate on the case of colour images captured in the visible spectrum.

It may seem natural to perform the processing on each channel individually. However, this does not
take into account the fact that the information gathered from each channel is not disjoint but has a certain, possibly
stronger, correlation across channels rather than within the space of each channel - in other words, the same pixel on each
channel most likely corresponds to the same object being photographed, whereas there's no such guarantee for any two
adjacent pixels in the same channel.\footnote{This depends on the field of interest. For example, in astronomy it may well
be the case that a channel of information corresponds to a specific wavelength of light, in which case it is not only true that
any one object is highly unlikely to appear across all wavelenghts, but the same absorption spectrum across different wavelenghts 
being captured at the same pixel most likely indicates two similar objects (e.g. gas clouds), one behind the other.}
Also, the problem remains of how to combine the resulting information, and this is not a straightforward
course of action in the case of segmentation.

Several methods to non-trivially process a colour image have previously been examined.
The structure-texture decomposition method suggested by Meyer \cite{Meyer_01} was recently extended to colour images
by Aujol and Kang \cite{Aujol_Kang_06} applying the G-norm to the RGB space.
Another interesting approach is presented for example in Tang, Sapiro and Caselles \cite{Tang_Sapiro_Caselles_01},
where the chromaticity values of each pixel
are mapped onto the unit sphere and then processed using a diffusion-based filter; the authors consider both isotropic
and anisotropic types. A rather similar approach seems to have been used in Yu and Bajaj \cite{Yu_Bajaj_02}, developed seemingly
independently. It was argued in Kang and Shen \cite{Chan_Kang_Shen_01} that a chromaticity-brightness
(CB) filter outperforms a hue-saturation-value (HSV) filter in combination with a total variation method.

\begin{figure}
\begin{center}
$\begin{array}{cc}
\multicolumn{1}{l}{\mbox{Image data}} &
\multicolumn{1}{l}{\mbox{Rounded composite}} \\
\epsfxsize=1.8in
\epsffile{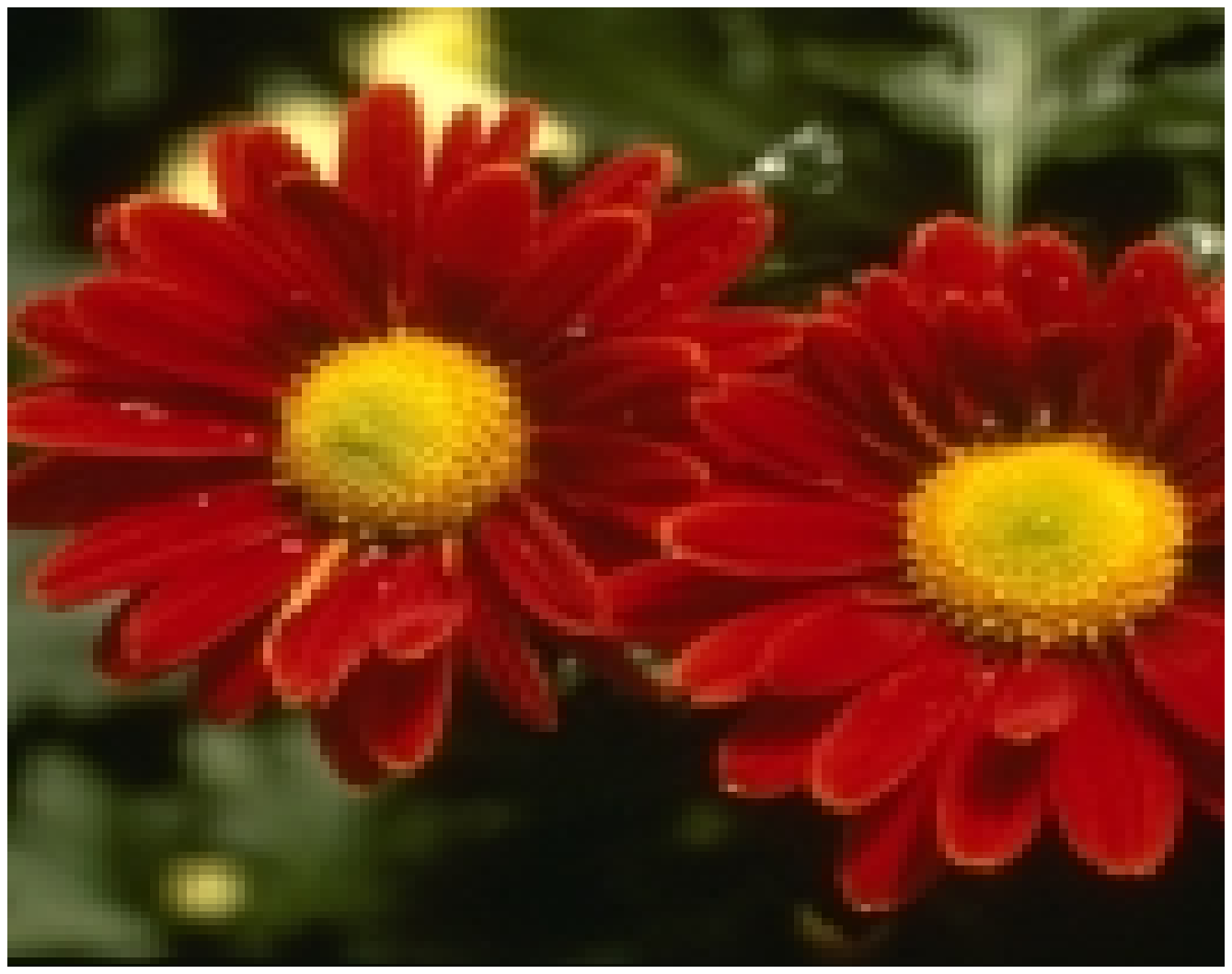} &
\epsfxsize=1.8in
\epsffile{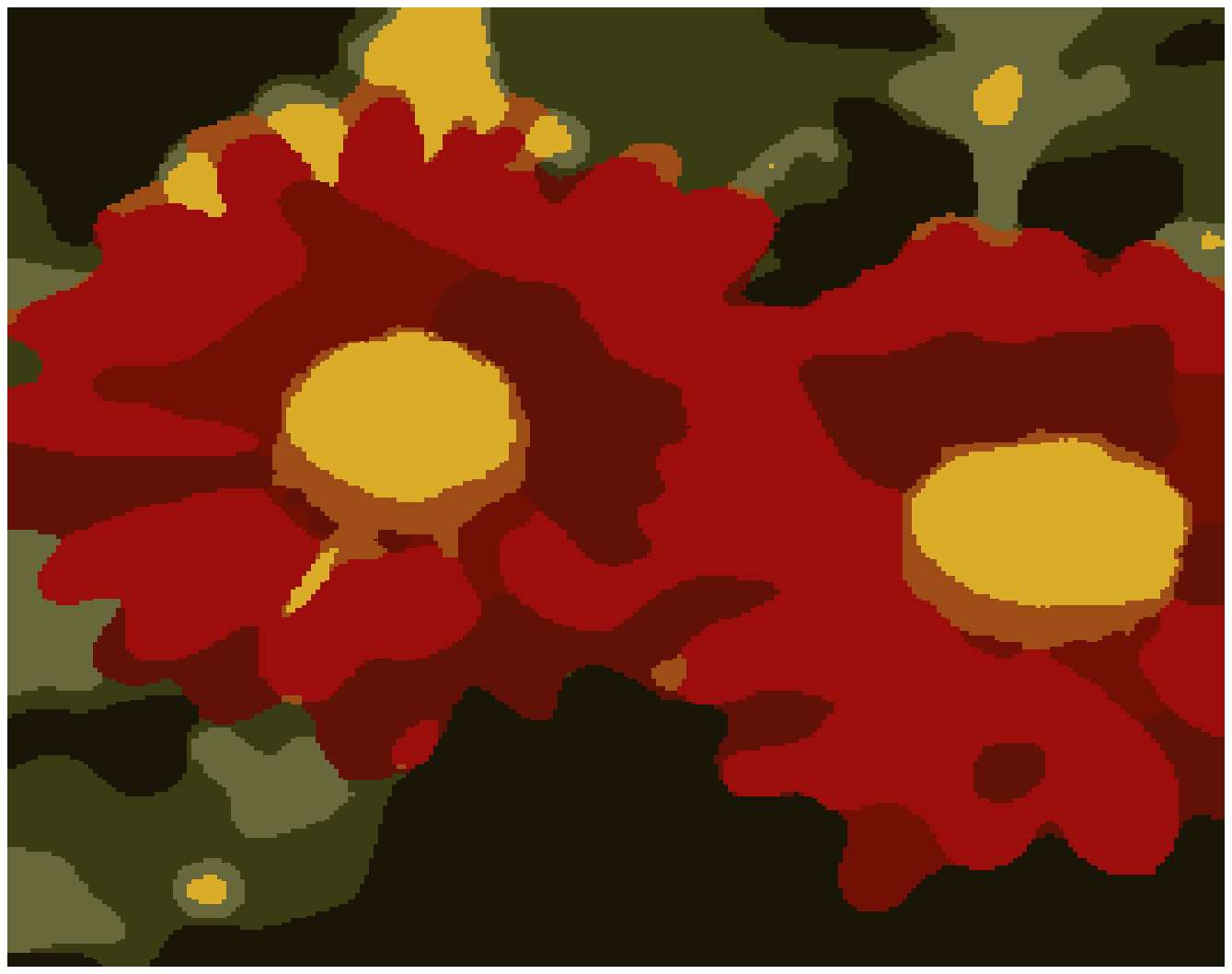} \\
\multicolumn{1}{l}{\mbox{Vectorial constituents}} &
\multicolumn{1}{l}{\mbox{Remainder}} \\
\epsfxsize=1.8in
\epsffile{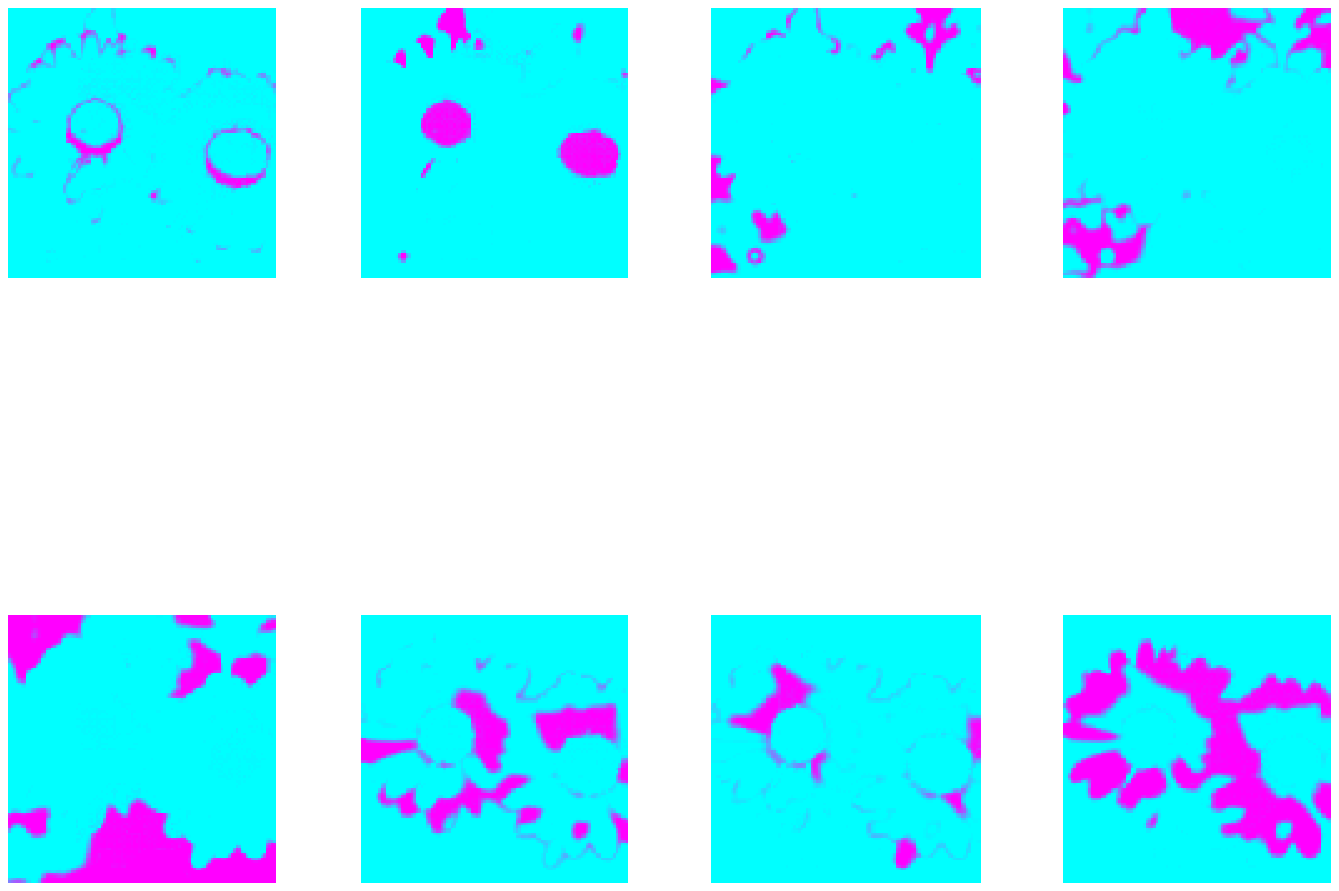} &
\epsfxsize=1.8in
\epsffile{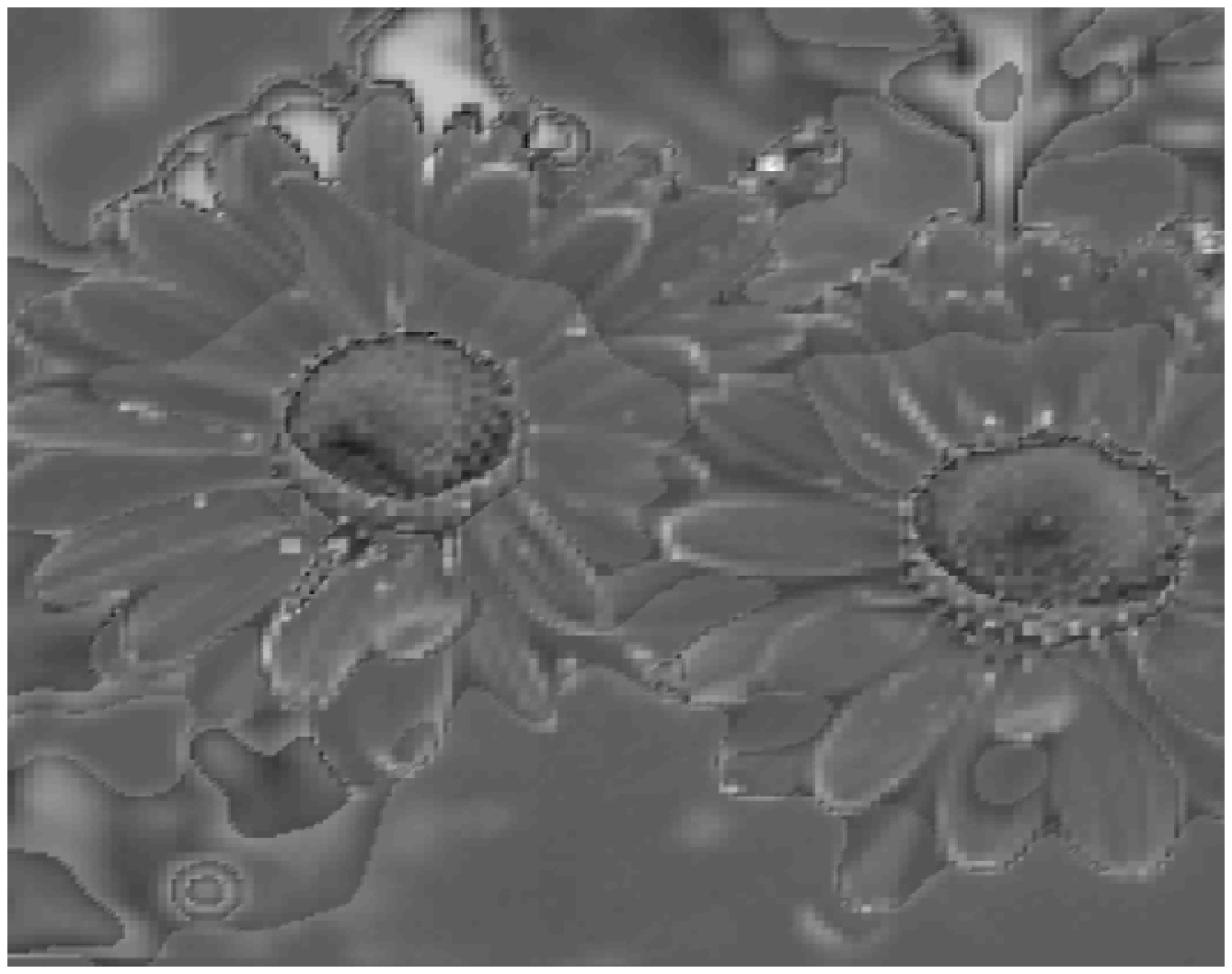} \\
\end{array}$
\end{center}
\caption{Flowers} 
\label{flowers_RGB}
\end{figure}

\begin{figure}
\begin{center}
$\begin{array}{cc}
\multicolumn{1}{c}{\mbox{Image data}} &
\multicolumn{1}{c}{\mbox{Rounded composite}} \\
\epsfxsize=1.8in
\epsffile{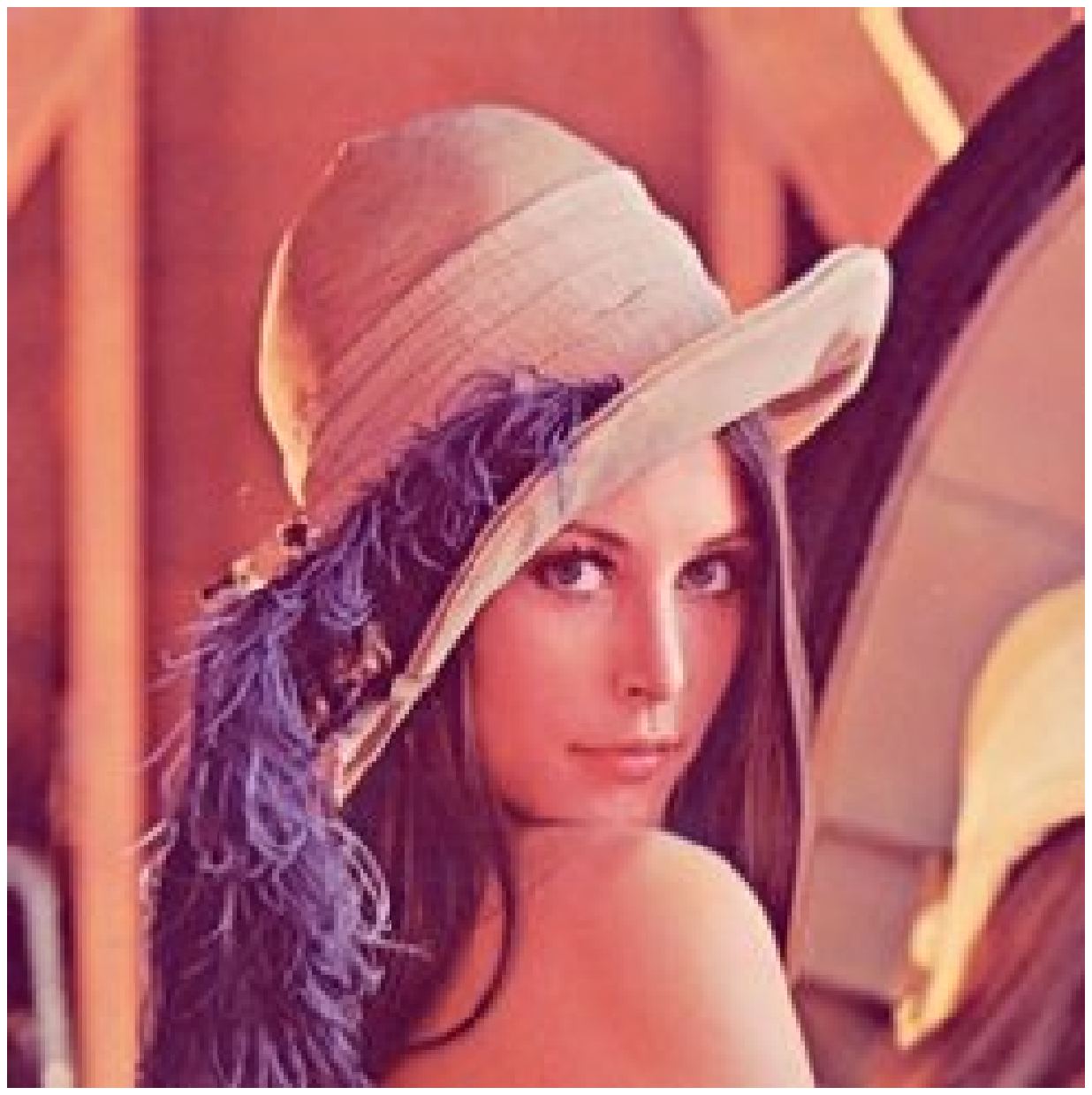} &
\epsfxsize=1.8in
\epsffile{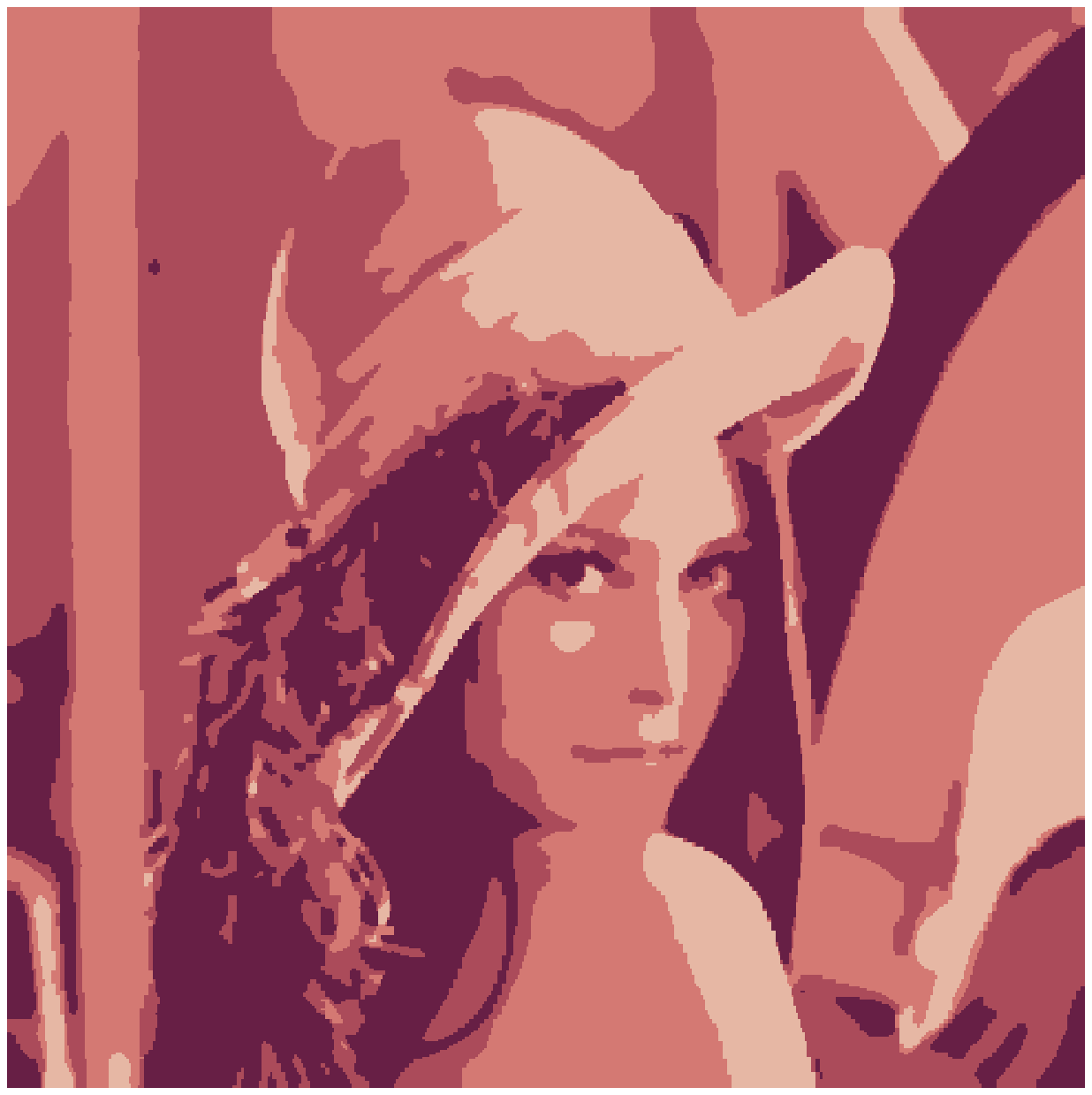} \\
\end{array}$
\end{center}
\caption{Lena}
\label{Lena_RGB} 
\end{figure}

We propose to use multi-channel information in our model by adapting the fitting terms in
(\ref{VV_fidelity}) to use the information from each of the colour channels $I_j$:
\[
c_{i,j}(u_i,I_j)=\frac{\int_{\Omega}u_i I_j } {\int_{\Omega}u_i}
\]
\[
F(c_{i,j},I_j) = \sum_{j=1}^{3} |I_j-c_{i,j}(u_i,I_j)|^2 ~dx \\
\]
This method can be followed regardless of how the multi-channel information is encoded, i.e. it can in principle
be applied to any colour space, and any number of channels. For simplicity, the RGB space has been used in computations so far.
Figures \ref{flowers_RGB} and \ref{Lena_RGB} were obtained using the RGB space. As can clearly be seen, the use of the RGB
space cannot distinguish between a difference in colour and one in luminosity, potentially leading to somewhat unrealistic segmentations
of complex real-world images; this could possibly be remedied by using the L*a*b* space and ignoring the luminosity component.

\section{Conclusion}

The method here presented is a flexible method for image segmentation and denoising based on the combination of
previously consolidated work. It can either be fine-tuned to specific goals or be allowed to run with almost no
user input other than the initial image, which can define its own natural scale. Thanks to the multigrid formulation,
computational costs will not spiral out of control. The method can also be further extended in several directions,
including adaptive mesh refinement and adaptive time-stepping to further reduce computational time; it can deal with
any desired colour space and any number of input and output channels, independent of each other.

The Esedo\={g}lu-Tsai formulation involving MBO thresholding has the advantage of dealing well with the well-known unstable
equilibrium value of $1/2$; however, we would suggest that there are important dynamics motivated by the Chan-Vese fitting
terms, especially at short time-scales, that are lost by the thresholding mechanism. Our multigrid solution
seeks to retain the computational swiftness while capturing the full dynamics of the equations.


\bibliography{thesisbib}
\bibliographystyle{abbrv}
\end{document}